\documentclass[10pt,journal,compsoc]{IEEEtran}
\usepackage[colorlinks,linkcolor=red]{hyperref}

\usepackage{algorithm}
\usepackage{algorithmic}
\usepackage{ragged2e}
\usepackage{amsmath}
\usepackage{amssymb}

\usepackage{color}
\usepackage{times}
\usepackage{epsfig}
\usepackage{graphicx}
\usepackage{url}
\usepackage{multirow}
\usepackage{booktabs}
\usepackage{subfig}

\usepackage{multirow}
\usepackage{booktabs}
\usepackage{amsthm}
\newtheorem{theorem}{Theorem}
\usepackage{textcomp}
\usepackage{stfloats}
\usepackage{url}
\usepackage{verbatim}
\usepackage{graphicx}
\usepackage{cite}

\usepackage[switch]{lineno}

\usepackage{xspace}
\makeatletter
\DeclareRobustCommand\onedot{\futurelet\@let@token\@onedot}
\def\@onedot{\ifx\@let@token.\else.\null\fi\xspace}
 
\def\ie{{i.e}\onedot}

\def\etal{{et al}\onedot}

\makeatother

\begin{document}

\title{TransFace++: Rethinking the Face Recognition Paradigm with a Focus on Accuracy, Efficiency, and Security}

\author{Jun Dan\textsuperscript{*}, Yang Liu\textsuperscript{*}, Baigui Sun, Jiankang Deng\textsuperscript{$\dagger$}, Shan Luo

\IEEEcompsocitemizethanks{
\IEEEcompsocthanksitem
\textsuperscript{*} Equal Contribution,
\textsuperscript{$\dagger$} Corresponding Author: Jiankang Deng. \protect
\IEEEcompsocthanksitem J. Dan is with Zhejiang University, HangZhou, China. Email: danjun@zju.edu.cn. \protect 
\IEEEcompsocthanksitem  B. Sun is with the Wolf 1069 b Lab, Sany Group, Guangzhou, China. Email: sunbaigui85@gmail.com. \protect 
 \IEEEcompsocthanksitem J. Deng is with the Department of Computing, Imperial College London, London, UK. E-mail: j.deng16@imperial.ac.uk. 
  \IEEEcompsocthanksitem Y. Liu and S. Luo are with the Department of Engineering, King’s College London, London WC2R 2LS, United
Kingdom. E-mails: \{yang.15.liu,
shan.luo\}@kcl.ac.uk.
 \protect
 }.}


\markboth{Journal of \LaTeX\ Class Files,~Vol.~14, No.~8, August~2015}%
{Shell \MakeLowercase{\textit{et al.}}: Bare Demo of IEEEtran.cls for Computer Society Journals}

\IEEEtitleabstractindextext{%
\begin{abstract}
\justifying  
Face Recognition (FR) technology has made significant strides with the emergence of deep learning. 
Typically, most existing FR models are built upon Convolutional Neural Networks (CNN) and take RGB face images as the model's input. 
In this work, we take a closer look at existing FR paradigms from high-efficiency, security, and precision perspectives, and identify the following three problems:
(i) CNN frameworks are vulnerable in capturing global facial features and modeling the correlations between local facial features.
(ii) Selecting RGB face images as the model's input greatly degrades the model's inference efficiency, increasing the extra computation costs.
(iii) In the real-world FR system that operates on RGB face images, the integrity of user privacy may be compromised if hackers successfully penetrate and gain access to the input of this model.
To solve these three issues, we propose two novel FR frameworks, \ie, TransFace and TransFace++, which successfully explore the feasibility of applying ViTs and image bytes to FR tasks, respectively.
Firstly, as revealed from our observations, we find that ViTs perform vulnerably when applied to FR scenarios with extremely large datasets.
We investigate the reasons for this phenomenon and discover that the existing data augmentation approaches and hard sample mining strategies are incompatible with ViTs-based FR backbone due to the lack of tailored consideration on preserving face structural information and leveraging each local token information.
To remedy these problems, we first propose a superior FR model called TransFace, which contains a patch-level data augmentation strategy named Dominant Patch Amplitude Perturbation (DPAP) and a hard sample mining strategy named Entropy-guided Hard Sample Mining (EHSM).
Furthermore, to improve inference efficiency and user privacy protection, we investigate the intrinsic property of image bytes and propose a superior FR model termed TransFace++.
The proposed model is trained directly on image bytes, presenting a novel approach to address the aforementioned issues.
Specifically, considering the importance of local correlations in bytes, an image bytes compression strategy named Topology-based Image Bytes Compression (TIBC) 
is introduced to extract prominent features from the raw bytes and integrate these features with byte embeddings, effectively mitigating information loss during the bytes mapping process.
Moreover, to strengthen the model's perception on geometric information encoded in image bytes, a novel cross-attention module named Structure Information-guided Cross-Attention (SICA) is designed to inject structure information into byte tokens for information interaction, significantly improving the model's generalization ability. 
Experiments on popular face benchmarks demonstrate the superiority of our TransFace and TransFace++.
Code is available at 
\url{https://github.com/DanJun6737/TransFace_pp}.
\end{abstract}

}

\maketitle

\IEEEdisplaynontitleabstractindextext

\IEEEpeerreviewmaketitle

\begin{figure}[t]
\begin{center}   
\includegraphics[width=0.8\linewidth]{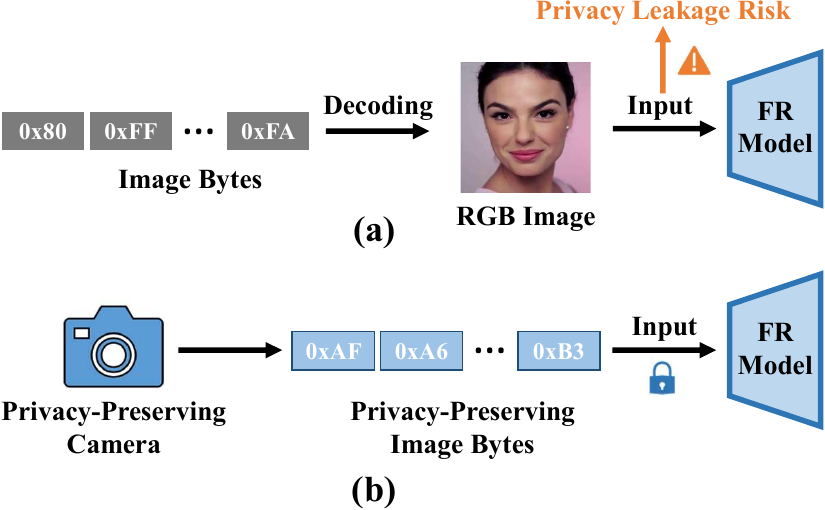}
\end{center}
\caption{
\textbf{(a):} Existing inference paradigms of FR models primarily rely on RGB face images, which poses a risk of privacy leakage for users.
\textbf{(b):} Our TransFace++ framework is able to directly operate on encrypted image bytes without reconstructing RGB images, greatly protecting user privacy.
}
\label{fig-camera}
\end{figure}

\IEEEraisesectionheading{\section{Introduction}
\label{intro}
}
\IEEEPARstart{W}{ith} the rise of deep learning, Face Recognition (FR) technology has made remarkable strides in real-world applications, including electronic payments, smartphone lock screens, and video surveillance.
However, most of the existing FR models are typically built upon Convolutional Neural Network (CNN) frameworks, and their inference paradigms primarily rely on RGB face images.
In this paper,
we delve into the landscape of prevailing FR paradigms from high efficiency, security, and precision perspectives, and identify the following three problems:
\textbf{(1) Lack of correlation between facial features:}
CNN-based FR models excel at extracting local facial features, but they have some limitations in capturing global facial features and modeling the correlations between different facial features within the same face image, which hampers the learning of fine-grained identity features and limits the improvement of the model's generalization performance.
\textbf{(2) Slow inference efficiency}:
Face images are typically stored as bytes on the disk. However, decoding bytes into RGB pixels can be highly time-consuming, 
which poses huge challenges to the CPU's computational head and slows down the data process significantly.
\textbf{(3) Poor security}:
In the realm of FR systems that operate on RGB face images, there is a potential risk to user privacy if hackers manage to breach and access the inputs to this model, as depicted in Fig.~\ref{fig-camera}. 

To understand the roots of these issues, we make several efforts to explore the transformer-based framework and how to substitute conventional RGB pixel representation with raw image byte data as the model's input. 
As an initial effort to address these issues, we conduct an in-depth quantitative and qualitative analysis to illustrate why the proposed methods are able to effectively resolve the aforementioned issues.
It should be emphasized that the objective of this research is to find a novel and efficient paradigm for the FR system instead of solving problems in a unified framework.
Thus, this has led us to propose two distinct frameworks: TransFace and TransFace++.
Some motivations and details of the proposed methods are explained as follows.

\textbf{Addressing Limitation (1).} When recognizing a person's identity, the human brain synthesizes and analyzes multiple facial signals (\emph{e.g.}, eyes, mouth, nose, and hairstyle) captured by the visual perception system to make the final decision~\cite{cornsweet2012visual}.
This decision-making process is highly similar to the self-attention mechanism in the Vision Transformer (ViT)~\cite{dosovitskiy2020image}, which aims to model the correlations between multiple patches and demonstrate powerful representation ability across various visual tasks. Motivated by the self-attention mechanism, in this paper, we aim to build a cutting-edge ViT-based FR framework called \textbf{TransFace} to address the first limitation.

Unlike CNN models, ViTs lack some convolution-like inductive biases, such as translation equivariance and locality, leading to challenges in the convergence of the ViTs backbone. 
To remedy this problem, pioneering works~\cite{touvron2021training,dosovitskiy2020image,chen2022transmix,zhou2022effective} point out that the rationale behind this derives from its data-hungry nature, indicating that a superior representation of ViTs would be supported by large-scale training data.
Taking advantage of their intrinsic data-hungry property, ViTs are commonly used to serve as an alternative backbone on various visual tasks~\cite{liu2021swin,dosovitskiy2020image,touvron2021training}.

\begin{figure}[t]
\begin{center}
   \includegraphics[width=0.8\linewidth]{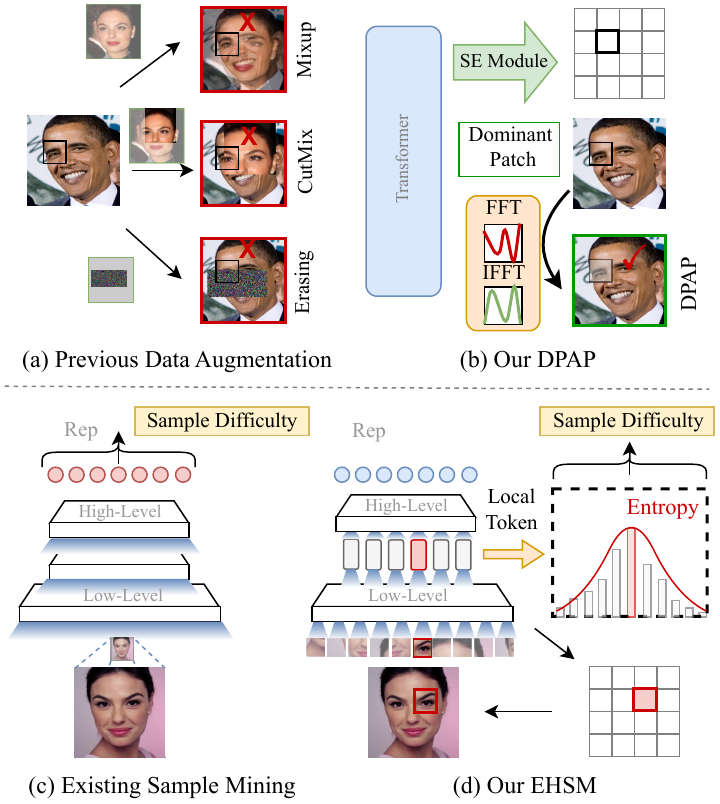}
\end{center}
\caption{
\textbf{Top:} Previous data augmentation approaches may destroy the fidelity and structural information of face identity when augmenting samples. Our DPAP strategy not only constructs diverse samples but also effectively preserves the key information of the face. \textbf{Bottom:} Existing hard sample mining methods usually adopt several instance-level indicators to measure sample difficulty, which is suboptimal for ViTs. Our EHSM strategy leverages information entropy from all local tokens to mine hard samples.}
\label{fig1}
\end{figure}

However, when considering an extremely data-adequate scenario to satisfy ViTs' data-hungry property, namely FR, we unexpectedly discover that the performance of ViTs is almost equal to that of CNNs~\cite{zhong2021face}. 
To explore why ViTs perform vulnerably in the FR realm, we investigate the ViT training process.
From a data-centric perspective, we discover that the instance-level data augmentation approach and the hard sample mining strategy are incompatible for the ViTs-based FR backbone due to the lack of tailored consideration of preserving the structural information of the face and using each local token information (illustrated in Fig. ~\ref{fig1}). 
To handle these drawbacks, we make the following two efforts: \textbf{(i)} Identify why and how to devise a patch-level data augmentation strategy on the ViT-based FR backbone.
\textbf{(ii)} Unveil why and how to extract representative token information. 
By comprehensive exploration and analysis in Sec.~\ref{sec:method}, we propose a superior FR model called TransFace, which
employs a patch-level data augmentation strategy named Dominant Patch Amplitude Perturbation (\textbf{DPAP}) and a hard sample mining strategy named Entropy-guided Hard Sample Mining (\textbf{EHSM}). Specifically, DPAP
randomly perturbs the amplitude information of dominant patches to expand sample diversity, effectively alleviating the overfitting
problem in ViTs. EHSM utilizes the information entropy in the local tokens to dynamically adjust the importance weight of the easy and hard
samples during training, leading to a more stable prediction.

\textbf{Addressing Limitations (2) and (3).}
To improve the model's efficiency and reduce CPU bottlenecks, some researchers have started focusing on training CNNs directly on image bytes (\emph{e.g.}, PNG and TIFF)~\cite{xu2020learning,ehrlich2019deep}. Unlike RGB pixels, image bytes are inherently sequential data, and therefore cannot be directly processed by conventional CNNs. Hence, these studies begin to modify the architectures of CNNs to accept bytes as input.
Due to the inherent advantage of the transformer in the processing of sequential data, recent works~\cite{park2023rgb,hortonbytes,yu2023megabyte} have shifted towards exploring
the direct training of ViTs in file bytes.

Although the aforementioned methods have been continuously suggested to modify the ViTs architecture for adapting bytes format, they still lack the quantitative analysis on the intrinsic property of image bytes. 
To investigate this, we first make several explorations in Sec.~\ref{sec:method}.
Based on this,  we find that existing methods inevitably destroy the local correlation of bytes and sever the connections between them,  causing the FR model to optimize under the incomplete direction.
To remedy this issue, based on the TransFace architecture, we propose a superior FR model named \textbf{TransFace++} that operates directly on image bytes. Notably, TransFace++ still adopts global attention operations to fully preserve the local correlation of bytes.

To alleviate the ViT's overall computational load caused by long sequential data, TransFace++ introduces a byte projector that maps input bytes into short-length byte embeddings, which greatly reduces the sequence length (from 37,600+ to 144).
However, we find that this sharp mapping process inevitably results in substantial information loss (\emph{e.g.}, some crucial facial features are lost), which has a detrimental effect on the subsequent FR task.
Motivated by signal compression theory~\cite{gersho2012vector,edelsbrunner2006persistence}, we introduce a Topology-based Image Bytes Compression (\textbf{TIBC}) strategy to solve this problem. 
Concretely, TIBC first models raw image bytes as a 1-dimensional time series signal and utilizes persistent homology (PH)~\cite{zomorodian2004computing} to separately track critical points (\emph{i.e.}, prominent features) and non-critical points (\emph{i.e.}, noise or redundant information) in the signal.
Then, the non-critical points are discarded, and only the critical points are retained to reconstruct the compressed signal (\emph{i.e.}, compressed bytes features).
Finally, compressed byte features are integrated with byte embeddings, which effectively mitigates information loss during the bytes mapping process.

Furthermore, unlike RGB face images, the image bytes are essentially sequential data and lack the geometric information presented in RGB images. Existing positional embedding techniques~\cite{dosovitskiy2020image,wu2021rethinking} are unable to effectively tackle this issue, which severely hinders the model's understanding of image bytes.
Inspired by the inherent invariance of the topological structure of the data~\cite{carlsson2009topology,moor2020topological}, we propose a novel Structure Information-guided Cross-Attention (\textbf{SICA}) module to extract the information of the underlying topological structure in image bytes. Notably, SICA also promotes the interaction between byte tokens and structure information, significantly improving the model's recognition performance.

Additionally, as illustrated in Fig.~\ref{fig-camera}, existing FR models rely heavily on RGB images for their inference paradigms. There is a significant risk of user privacy leakage if hackers illegally obtain the model's input. 
It is worth mentioning that the proposed TransFace++ framework holds great potential for application in privacy-preserving FR tasks. 
Consider a privacy-preserving camera that can convert captured face images into encrypted image bytes. 
Our TransFace++ model is capable of performing identity recognition directly on the encrypted image bytes without reconstructing the RGB images, effectively protecting user privacy.

The following are the main contributions of this paper.

(1) We propose a superior ViT-based FR backbone called TransFace which employs a patch-level data augmentation strategy DPAP to alleviate the overfitting issue in ViTs and a novel hard sample mining strategy EHSM to enhance the stability of FR model prediction.

(2) We propose a cutting-edge FR backbone named TransFace++ that operates directly on images bytes. An image bytes compression strategy TIBC is introduced to mitigate information loss, and a novel cross-attention module SICA is devised to inject structure information into image bytes.

(3) To the best of our knowledge, we are the first to explore directly building the FR model on large-scale image bytes dataset, which opens up a viable path for future research on FR privacy-preserving systems. Experimental results have indicated the superiority of TransFace++ over existing privacy-preserving FR methods.

(4) Extensive experiments and analysis on various facial benchmarks have shown the superior recognition performance of our TransFace and TransFace++.

The preliminary results of this work have been published in~\cite{dan2023transface}. The current work has been improved and extended from the conference version in the following aspects.
(i) We propose a superior FR framework termed TransFace++ that operates directly on image bytes. It utilizes an image bytes compression strategy named TIBC to minimize information loss during the bytes mapping process, along with a novel cross-attention module named SICA to strengthen the model’s perception on geometric information encoded in image bytes.
(ii) More importantly, our TransFace++ can be effectively combined with privacy-preserving cameras, allowing for direct inference on encrypted image bytes without the necessity of reconstructing RGB images. This FR paradigm greatly protects user privacy. The experimental results on the privacy-preserving FR task show that our proposal works better than existing privacy-preserving FR methods, validating the superiority of our method. To the best of our knowledge, we are the first to explore directly building FR models on large-scale image bytes dataset, which opens up a viable path for future research on FR privacy-preserving systems.
(iii) The experimental results on multiple benchmark datasets indicate that our TransFace++ models even surpass some FR competitors trained with RGB tensors, implying the effectiveness of our method. Notably, unlike FR models that perform inference on RGB tensors, our TransFace++ framework directly makes predictions on image bytes, bypassing the complex image decoding steps, and thereby greatly boosting the real-time capability of FR systems.

\section{Related Works}

\textbf{Vision Transformer (ViT).}
Recently, ViT has demonstrated its powerful feature representation ability in various vision tasks, including image recognition~\cite{dosovitskiy2020image,touvron2021training,liu2021swin}, semantic segmentation~\cite{li2024transformer}, and object detection~\cite{hao2025simple}.
Unlike CNN, ViT is mainly based on the self-attention mechanism, which can effectively capture the relationships between different tokens. It has been demonstrated that ViT does not generalize well when trained on insufficient amounts of training samples~\cite{dosovitskiy2020image}. 

To improve the performance of ViTs, previous works~\cite{kim2024learning,han2021transformer} mainly try to modify the structure of the models, which is very dependent on experience.
For instance, DeiT~\cite{touvron2021training} introduces a novel knowledge distillation procedure to learn discriminative features. TNT~\cite{han2021transformer} employs an inner
transformer block to model the fine-grained relationship among sub-patches. 
Recently, several works have been proposed to strengthen the generalization ability of ViTs from a data-centric perspective. 
Touvron \etal~\cite{touvron2021training} show that strong data augmentation strategies can help ViTs absorb data more efficiently.
TransMix~\cite{chen2022transmix} introduces a
data augmentation strategy that leverages the transformer's attention maps to guide the mixture of labels. TokenMix~\cite{liu2022tokenmix} proposes to mix images at the token level, which facilitates ViTs to infer sample classes more accurately.

\noindent
\textbf{Face Recognition (FR).}
CNNs have made significant progress in face-related tasks.
Among them, extracting deep face embedding attracts the attention of many researchers.
There are two main ways to train CNNs for FR.
One kind of way is metric learning-based methods, which aim to learn discriminative face representation, such as Triplet loss~\cite{schroff2015facenet}, Tuplet loss~\cite{sohn2016improved} and Center loss~\cite{wen2016discriminative}.
Another way is to use margin-based softmax methods, which focus on incorporating the margin penalty into the softmax classification loss framework, such as ArcFace~\cite{deng2019arcface}, CosFace~\cite{wang2018cosface}, UniFace~\cite{zhou2023uniface}, and MagFace~\cite{meng2021magface}. To further boost the model's accuracy on the large-scale dataset, some studies' emphasis has shifted
to data uncertainty~\cite{li2021spherical}, adaptive parameters~\cite{kim2022adaface,meng2021magface}, inter-class regularization~\cite{duan2019uniformface}, sample mining~\cite{huang2020curricularface, dan2024topofr}, learning acceleration~\cite{an2021partial,an2022killing}, knowledge distillation~\cite{caldeira2024mst,boutros2024adadistill}, etc.

The recently proposed face transformer~\cite{zhong2021face} first demonstrates the feasibility of using ViT in FR. However, there is still a lack of exploration on how to train a superior ViT-based FR model on the extremely large-scale dataset~\cite{qin2023swinface}.

\noindent
\textbf{Training Models Directly on File Bytes.}
Recently, some studies~\cite{gueguen2018faster, ehrlich2019deep} have attempted to skip the image decoding steps to improve CNNs' efficiency.
Specially, Gueguen \etal~\cite{gueguen2018faster} 
utilize features extracted from
Discrete Cosine Transform (DCT) coefficients to classify images, effectively reducing decoding overhead.
Xu \etal~\cite{xu2020learning} adapt the CNN architecture to accept DCT coefficients as input, preventing information loss from downsampling.
Moreover, recent works~\cite{yu2023megabyte,hortonbytes,park2023rgb} have started to explore the training of ViT using file bytes.
For example, Park and Johnson~\cite{park2023rgb} have indicated that training ViT from the DCT coefficients of JPEG can greatly improve its speed. ByteFormer~\cite{hortonbytes} has shown the potential to classify images and audio directly from bytes. However, these methods inevitably destroy the local correlation of file bytes.

\section{Background of TransFace++}
In this section, we will briefly introduce some key concepts of Persistent Homology (PH) in Sec.~\ref{PH}, and common image byte encodings in Sec.~\ref{image-bytes}. More details of the Persistence Diagram (PD) and Persistence Image (PI) can be found in~\cite{adams2017persistence,zomorodian2004computing}. 

\subsection{Persistent Homology (PH)}
\label{PH}
PH is a computational topology technique that captures the changes in the topological invariants of a simplicial complex as the scale parameter $\epsilon$ is varied.

\noindent
\textbf{Notation.}
$\mathcal{X}:=\left \{x_{i} \right \}_{i=1}^{n}$ denotes a point cloud and $\phi:\mathcal{X}\times\mathcal{X}\rightarrow \mathbb{R}$ denotes
a distance metric on $\mathcal{X}$. 

\noindent
\textbf{Vietoris-Rips (VR) Complex.}
The VR complex~\cite{zomorodian2010fast} is a unique simplicial complex constructed from a set of points in a metric space, with the purpose of approximating the topology of the underlying space.
For $0\leq \epsilon< +\infty$, the VR complex of the point cloud $\mathcal{X}$ on scale $\epsilon$ is denoted as $\mathcal{VR}_{\epsilon}(\mathcal{X})$. It contains all the simplices (\emph{i.e.}, subsets) of $\mathcal{X}$, and each component of the point cloud $\mathcal{X}$ adheres to the following distance constraint:
$\phi(x_{i},x_{j})\leq \epsilon$ for any $i,j$. 
Additionally, the VR complex satisfies a nesting rule, \emph{i.e.}, $\mathcal{VR}_{\epsilon_{i}}\subseteq \mathcal{VR}_{\epsilon_{j}}$ for any $\epsilon_{i}\leq\epsilon_{j}$, which enables us to track the progressive evolution of the simplical complex as the scale $\epsilon$ increases.

\noindent
\textbf{Homology Group.}
The homology group is an algebraic structure that analyzes the topological features of a simplicial complex in various dimensions $\gamma$, including connected components ($\gamma=0$), cycles/loops ($\gamma=1$), voids/cavities ($\gamma=2$), and higher-dimensional features ($\gamma\geq3$). 
By tracking the changes in the homology group with increasing scale $\epsilon$, valuable insights can be gained regarding the multi-scale topological information of the underlying space.

\noindent
\textbf{Persistence Diagram (PD).}
PD $\mathrm{D}_{\gamma}$ is a multi-set of points $(\varpi, \tau)$ in the Cartesian plane $\mathbb{R}^{2}$, which encodes the birth time $\varpi$ and death time $\tau$ of each topological feature.
Concretely, each persistence point $(\varpi, \tau)$ corresponds to a topological feature that emerges at scale $\varpi$ and disappears at scale $\tau$.
The lifetime of the homology group is defined as the difference between the death and the birth time, \emph{i.e.}, $l_{f} = \left | \varpi-\tau\right |$.
However, since each PD is a multi-set of points and
cannot be directly processed by the modern network frameworks~\cite{som2018perturbation}.
Therefore, it is necessary to vectorize them before using them as input.

\noindent
\textbf{Persistence Image (PI).}
PI $\mathrm{I}$~\cite{adams2017persistence} is a finite-dimensional vector representation of a PD. It can be obtained through the following steps: (1) First, a PD is initially mapped to an integrable function $\rho_{I}:\mathbb{R}^{2}\rightarrow\mathbb{R}$, referred to as a persistence surface. This surface $\rho_{I}$ is defined as a weighted sum of Gaussian functions, each function centered at a point in the PD.
(2) Next, a grid is formed by discretizing a subdomain of the persistence surface $\rho_{I}$.
(3) Finally, PI $\mathrm{I}$ is created by integrating the persistence surface $\rho_{I}$ into each grid box, resulting in a matrix of pixel values.

\begin{figure*}[t]
\begin{center}
\includegraphics[width=0.8\linewidth]{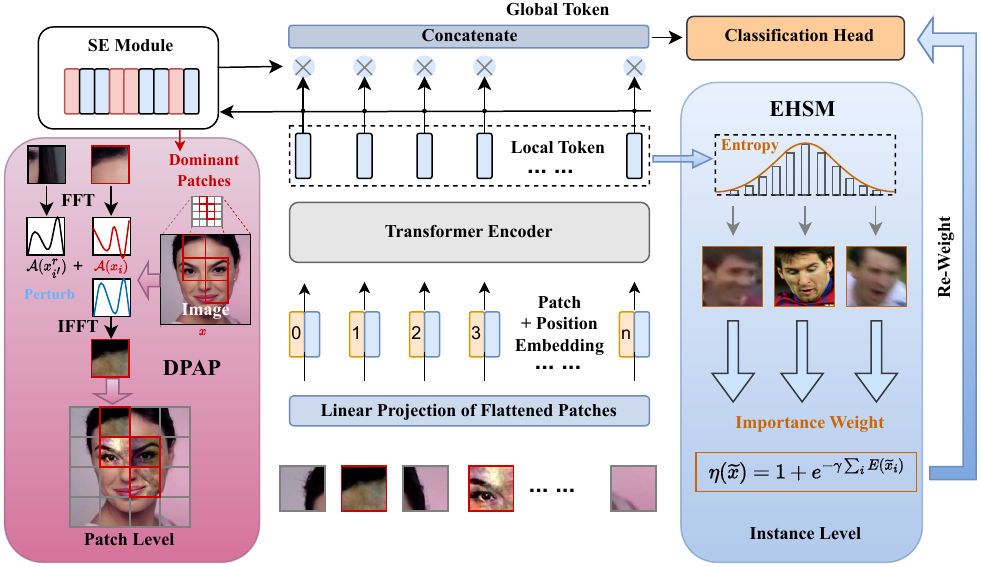}
\end{center}
   \caption{Global overview of the proposed TransFace model. To alleviate the overfitting problem in ViTs, the DPAP strategy employs the SE module to screen out the top-$K_{0}$ dominant patches, and then randomly perturbs their amplitude information to expand sample diversity. 
   Furthermore, to effectively mine hard samples and enhance the feature presentation power of local tokens, the EHSM strategy utilizes an entropy-aware weight mechanism to re-weight the classification loss.
    $n$ is the total number of patches, and $\bigotimes$ denotes the multiplication operation between the local token and the scaling factor generated by the SE module. The image patches with red boxes represent dominant patches.}
\label{fig-transface}
\end{figure*}

\subsection{Common Image Byte Encodings}
\label{image-bytes}
\noindent
\textbf{fCHW:}
The notation fCHW represents flattened tensors arranged in channel, height, and width order. It specifically denotes uint8 image bytes stored in CHW order, without any accompanying file headers. Storing images in this manner is uncommon, as it necessitates prior knowledge of their height and width for successful decoding.
However, this format can serve as a baseline for comparison and analysis.

\noindent
\textbf{fHWC:}
This format is similar to fCHW, but image bytes are
stored in ``HWC" order.

\noindent
\textbf{TIFF:}
The TIFF byte encoding offers various custom configurations.
We adopt the default settings provided by PIL~\cite{clark2015pillow}. This format includes TIFF image headers that provide details about configuration options and image size.

\noindent
\textbf{PNG:}
The PNG format~\cite{boutell1997png} includes PNG configuration headers and rows of image data stored in ``IDAT" chunks. Each chunk contains a byte indicating the filtering method applied to that row's data.
The filtering method uses neighboring pixel values to determine offsets for the row's data.
Our PNG file consists of RGB data rows with applied offsets, occasionally interrupted by bytes for file encoding settings. We do not utilize PNG's optional zlib compression.

\section{Proposed Approach}
\label{sec:method}

\subsection{TransFace}

\subsubsection{Preliminaries}
Different from the original ViT architecture, we do not employ the learnable class token in our model.  
The architecture of our TransFace model is depicted in Fig.~\ref{fig-transface}. It is made up of four components: a linear projection layer $\Psi$, a transformer encoder $\mathcal{F}$, a ``Squeeze-and-Excitation" (SE) module $\mathcal{S}$~\cite{8578843} and a classification head $\mathcal{C}$. We will detail the architecture of TransFace in Sec.~\ref{DPAP}.

\subsubsection{Motivation}
Here, we first introduce the motivation behind the proposed TransFace from the perspective of patch-level data augmentation strategy and hard sample mining strategy, respectively.

\noindent\textbf{Patch-level Data Augmentation Strategy.} Due to the lack of inductive biases, ViT-based models are hard to train and are prone to overfitting~\cite{touvron2021training,dosovitskiy2020image,chen2022transmix}.
To alleviate the overfitting phenomenon, existing works~\cite{touvron2021training,zhou2022effective} attempt several data augmentation strategies, such as Random Erasing~\cite{zhong2020random}, Mixup~\cite{zhang2017mixup}, CutMix~\cite{yun2019cutmix}, RandAugment~\cite{cubuk2020randaugment} and their variants~\cite{chen2022transmix,touvron2021training,liu2022tokenmix}, to construct diverse training samples. 
However, these instance-level data augmentation strategies are not suitable for the FR task because they will inevitably destroy some key structural information about face identity (as shown in the top of Fig.~\ref{fig1}), which may lead ViTs to optimize in the incorrect direction.
Furthermore, a recent study~\cite{zhou2022effective} observed that ViTs are prone to overfitting to certain local patches during training, resulting in severely impaired generalization performance of the model.
For instance, in the FR task, the prediction of ViT may be dominated by a few facial patches (\emph{e.g.}, eyes, and forehead). Therefore, once these key patches are disturbed (\emph{e.g.}, a superstar wearing sunglasses or a hat), the model tends to make spurious decisions. 
These issues seriously affect the large-scale deployment of ViT-based FR models in real scenarios.

To solve the aforementioned problem, we introduce a patch-level data augmentation strategy named Dominant Patch Amplitude Perturbation (\textbf{DPAP}).
Without destroying the fidelity and structural information of the face, DPAP can efficiently expand sample diversity.
Concretely, DPAP uses a Squeeze-and-Excitation (SE) module~\cite{8578843} to select the top-$K_{0}$ patches (\textbf{dominant patches}), then randomly mixes their amplitude information and combines it with the original phase information to generate diverse samples.

\noindent\textbf{Hard Sample Mining Strategy.} As demonstrated in~\cite{lin2017focal,huang2020improving}, hard sample mining technology plays an important role in boosting the model's final performance by continuously assimilating knowledge from effective/hard samples. 
Most previous works are specially designed for CNNs, they usually adopt several instance-level indicators of the sample, such as prediction probability~\cite{lin2017focal,huang2020curricularface}, prediction loss~\cite{fan2017learning,shrivastava2016training}, and latent features~\cite{schroff2015facenet}, to mine hard samples (as shown at the bottom of Fig.~\ref{fig1}). However, a recent study~\cite{zhou2022effective} has shown that the ViT prediction is mainly determined by only a few patch tokens, which means that the global token of ViTs may be dominated by a few local tokens. 
Therefore, directly using such biased indicators to extract hard samples is suboptimal for ViTs. 

To better mine hard samples, we propose a 
novel hard sample mining strategy named Entropy-guided Hard Sample Mining (\textbf{EHSM}). EHSM treats the ViT as an information processing system, which dynamically adjusts the
importance weight of easy and hard samples in terms of the total amount of information contained in the local tokens.
It is worth mentioning that EHSM has the potential to encourage ViT to fully mine fine-grained information contained in each face patch, especially some less-attended facial cues (\emph{e.g.}, lip and jaw), which greatly enhances the feature representation power of each local token (as verified by our experiment in Fig.~\ref{fig-ehsm}). In this way, even if some important image patches are destroyed, the model can also make full use of the remaining facial cues to generalize the global token, leading to a more stable prediction.

\subsubsection{Patch-Level Data Augmentation Strategy}
\label{DPAP}
As analysed above, existing works mainly apply a series of instance-level data augmentation strategies to alleviate the overfitting phenomenon in ViTs.
However, for FR tasks, these strategies will inevitably destroy some key structural information of face identity (as shown in Fig.~\ref{fig1}), which may seriously affect the learning of discriminative face tokens.
Furthermore, a recent study~\cite{zhou2022effective} observed that ViTs are prone to overfitting to a few patches, greatly restricting the deployment and scalability of ViT-based FR models in application scenarios.

\begin{figure}[t]
\begin{center}   
\includegraphics[width=0.8\linewidth]{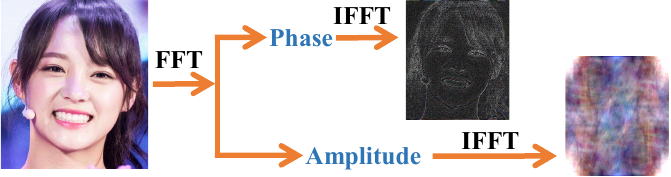}
\end{center}
\caption{
Demonstration of phase-only reconstructed face image and amplitude-only reconstructed face image.
\textbf{Upper row:} Face image is reconstructed using only the phase information by setting the amplitude information to a constant. \textbf{Bottom row:} Face image is reconstructed using only the amplitude information by making the phase component constant.
}
\label{fig-phase}
\end{figure}

To address the aforementioned issues, a novel patch-level data augmentation strategy named Dominant Patch Amplitude Perturbation (\textbf{DPAP}) is proposed for the ViTs-based FR backbone.
The basic motivation for DPAP arises from a well-known property of Fourier transformation~\cite{piotrowski1982demonstration}: the
Fourier phase spectrum preserves the high-level semantics of the original signal, while the Fourier amplitude spectrum embodies low-level statistics.
Specifically, for a face image, we find that the amplitude spectrum encodes style information, and the phase spectrum captures structural and semantic information of face identity, as shown in Fig.~\ref{fig-phase}.
Therefore, mixing the amplitude spectrum while preserving the phase helps to generate diverse face images and retain crucial information of face identity.

The main steps of the DPAP strategy are:
(1) First, we insert an SE module $\mathcal{S}$ into the output of the transformer encoder $\mathcal{F}$, and utilize the scaling factors generated by $\mathcal{S}$ to find out the top-$K_{0}$ patches (\emph{i.e.}, \textbf{dominant patches}) of the original image $x$ 
that contribute the most to the final prediction. (2) Second, we employ a linear mixing mechanism to randomly perturb the amplitude information of these dominant patches. (3) Finally, we feed the reconstructed image $\widetilde{x}$ into our TransFace model for supervised training.

Mathematically, given an image $x$, we denote a sequence of decomposed image patches as $x=(x_{1},x_{2},\cdots,x_{n})$, where each $x_{i}$ represents an image patch and $n$ is the total number of patches.
And the output of the transformer encoder $\mathcal{F}$ is denoted as $( f_{1},f_{2},\cdots ,f_{n})$, where $f_{1},\cdots ,f_{n}$ represent the local tokens. 
Then, all the local tokens extracted by $\mathcal{F}$ will pass through the SE module $\mathcal{S}$ and be re-scaled as $( \kappa_{1}\cdot f_{1},\kappa_{2}\cdot f_{2},\cdots ,\kappa_{n}\cdot f_{n})$, where $\kappa_{1}, \cdots, \kappa_{n}$ denote the scaling factors generated by $\mathcal{S}$.
These scaling factors $\kappa_{1},\cdots,\kappa_{n}$ of local tokens indirectly reflect the importance of local tokens in prediction. We further normalize these scaling factors using the softmax function:
\begin{equation}
d_{i}=\frac{e^{\kappa_{i}}}{\sum_{i'}e^{\kappa_{i'}}}.
\end{equation}

According to the top-$K_{0}$ largest normalized scaling factors, we can select the top-$K_{0}$ dominant patches that the model ``cares about" the most. 
To relieve ViTs from overfitting to these dominant patches, a natural idea is to let the model ``see'' more diverse patches.

Motivated by the structural information preservation property of the Fourier phase spectrum~\cite{piotrowski1982demonstration,yang2020fda}, 
we propose to perturb the amplitude spectrum information of their dominant patches using a linear mix mechanism and keep their phase spectrum information unchanged.

For a single-channel image patch $x_{i}$, its Fourier transform $\mathcal{T}(x_{i})$ can be expressed as:
\begin{equation}
\mathcal{T}(x_{i})(u,v)=\sum_{h=0}^{H-1}\sum_{w=0}^{W-1}x_{i}(h,w)e^{-j2\pi(\frac{h}{H}u+\frac{w}{W}v)}, 
\end{equation}
where $j^{2}=-1$, $H$ and $W$ represent the height and width of $x_{i}$, respectively. $\mathcal{T}^{-1}(x_{i})$ represents the corresponding inverse Fourier transform that maps the amplitude and phase information back to the original image space. The Fourier transform and its inverse transform can be efficiently implemented by the FFT and IFFT algorithms~\cite{brigham1967fast}, respectively. 

$\mathcal{A}(x_{i})$ and
$\mathcal{P}(x_{i})$ represent the amplitude spectrum and phase spectrum of the image patch $x_{i}$, respectively:
\begin{equation}
\begin{split}
&\mathcal{A}(x_{i})(u,v)=\left [ \mathcal{R}^{2}(x_{i})(u,v) + \mathcal{I}^{2}(x_{i})(u,v) \right ]^{1/2}, \\
&\mathcal{P}(x_{i})
(u,v)=\arctan \left [ \frac{ \mathcal{I}(x_{i})(u,v) }{\mathcal{R}(x_{i})(u,v) }\right ], \\ 
\end{split}
\end{equation}
where $\mathcal{R}(x_{i})$ and $\mathcal{I}(x_{i})$ denote the real and imaginary parts of $\mathcal{T}(x_{i})$, respectively.
For RGB face image patches, we need to calculate the Fourier transform of each channel independently to obtain the final amplitude spectrum and phase spectrum.
 
To effectively construct diverse images without destroying the fidelity and structural information of face identity, we employ a mix-up mechanism to linearly mix the amplitude spectrum of dominant patch $x_{i}$ and random patch $x^{r}_{i'}$:
\begin{equation}
\widetilde{\mathcal{A}}(x_{i})=\lambda{\mathcal{A}}(x_{i})+(1-\lambda){\mathcal{A}}(x^{r}_{i'}),
\end{equation}
where $\lambda \sim U(0,\alpha)$, $U(0,\alpha)$ is a uniform distribution on $\left [ 0,\alpha\right ]$, 
$x^{r}_{i'}$ is a random patch of a random training sample $x^{r}$,
and $\alpha$ is a hyper-parameter used to control the intensity of amplitude information mixing.
We then combine the mixed-amplitude spectrum with the original phase spectrum to reconstruct a new Fourier representation:
\begin{equation}
\mathcal{T}(\widetilde{x_{i}})(u,v)=\widetilde{\mathcal{A}}(x_{i})(u,v)*e^{-j*\mathcal{P}(x_{i})(u,v)},
\end{equation}
which will be mapped to the original image space by the inverse Fourier transform to generate a new patch, \emph{i.e.}, $\widetilde{x_{i}}=\mathcal{T}^{-1}\left [\mathcal{T}(\widetilde{x}_{i})(u,v) \right ]$. 
We then feed the augmented image $\widetilde{x}$ into the model for supervised training. 

For the augmented image $\widetilde{x}$, we denote the local tokens extracted by $\mathcal{F}$ as $( \widetilde{f}_{1},\widetilde{f}_{2},\cdots ,\widetilde{f}_{n})$,
which will be further fed into the SE module $\mathcal{S}$ and re-scaled as $( \widetilde{\kappa}_{1}\cdot \widetilde{f}_{1},\widetilde{\kappa}_{2}\cdot \widetilde{f}_{2},\cdots ,\widetilde{\kappa}_{n}\cdot \widetilde{f}_{n})$. Then, all the rescaled local tokens are concatenated into a global token $\widetilde{g} =\left [ \widetilde{\kappa}_{1}\cdot \widetilde{f}_{1};\widetilde{\kappa}_{2}\cdot \widetilde{f}_{2};\cdots; \widetilde{\kappa}_{n}\cdot \widetilde{f}_{n}\right ]$, which will be used for the subsequent FR task via the classification head $\mathcal{C}$.
In our model, we adopt the most widely used ArcFace Loss~\cite{deng2019arcface} as the basic classification loss:
\begin{equation}
\mathcal{L}_{arc}(\widetilde{g},y)=-\log\frac{e^{s(\cos(\theta_{y}+m))}}{e^{s(\cos(\theta_{y}+m))}+\sum_{l=1,l\neq y}^{c}e^{s\cos\theta_{l}}},
\end{equation}
where $y$ denotes the class label of image $x$,
$s$ is a scaling hyperparameter, $\theta_{l}$ is the angle between 
the $l$-th column weight 
and feature, $m>0$ denotes an additive angular margin and $c$ is the class number.

As the dominant patches are constantly changed, the DPAP strategy can indirectly encourage the FR model to use other facial patches, especially some patches that are easily ignored by the deep network (\emph{e.g.}, ears, mouth and nose), to
make a more confident prediction.
More importantly, different from previous data augmentation strategies, the DPAP strategy cleverly utilizes the prior knowledge provided by the model (\emph{i.e.}, the position of the dominant patches) to augment the data, which can more precisely mitigate the problem of overfitting and improve the generalizability of ViTs.

\subsubsection{Entropy-based Hard Sample Mining Strategy} 
In~\cite{lin2017focal,huang2020improving}, hard sample mining technology is often adopted to further boost the final performance of the model. Previous works on hard sample mining, such as Focal loss~\cite{lin2017focal}, MV-Softmax~\cite{wang2020mis}, OHEM~\cite{shrivastava2016training}, AT$_{k}$ loss~\cite{fan2017learning}, etc., are specially designed for CNNs, they aim to encourage models to explicitly emphasize the impact of hard samples~\cite{huang2020curricularface,schroff2015facenet}.
These methods usually utilize several instance-level indicators of the sample, such as prediction probability~\cite{lin2017focal,huang2020curricularface}, prediction loss~\cite{fan2017learning,shrivastava2016training}, and latent features~\cite{schroff2015facenet}, to directly or indirectly measure sample difficulty.
However, a recent study~\cite{zhou2022effective} has shown that the ViT prediction is mainly determined by only a few patch tokens, which means that the global token of ViT can be dominated by several local tokens. Therefore, for ViTs, it is suboptimal to use such biased indicators to mine hard samples.

\begin{figure}[t]
\begin{center}   \includegraphics[width=1.0\linewidth]{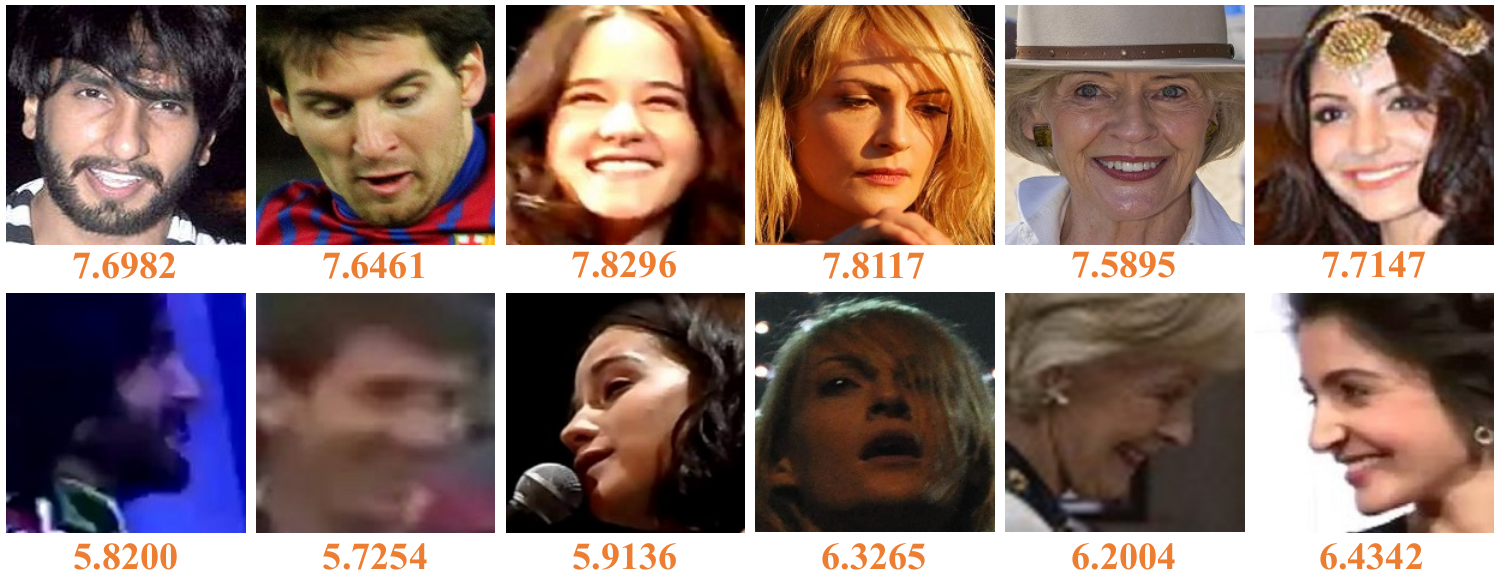}
\end{center}
   \caption{Example images and corresponding information entropy. Samples labeled with the same ID are displayed in each column. \textbf{First row:} Easy samples usually contain richer information (\emph{i.e.}, larger information entropy). \textbf{Second row:} Hard samples usually contain less information (\emph{i.e.}, lower information entropy).
   }
\label{fig-entropy}
\end{figure}

To mine hard samples more precisely, motivated by the information theory~\cite{cover1999elements}, we propose to measure sample difficulty in terms of the total amount of information contained in the local tokens. As demonstrated in Fig.~\ref{fig-entropy}, high-quality face images (easy samples) usually contain richer information (\emph{i.e.}, higher information entropy) and therefore are easier to learn by the model. Low-quality face images (hard samples), such as blurred face images and low-contrast face images, usually contain relatively less useful information (\emph{i.e.}, lower information entropy), so they are more difficult to learn.  

When we consider a deep neural network $\mathcal{M}$ as an information processing system, we can abstract its topology into a graph $\mathcal{G}=(\mathcal{Z},\mathcal{Q})$. A series of neurons form the vertex set $\mathcal{Z}$, and the connections between neurons form the edge set $\mathcal{Q}$. For any $z\in\mathcal{Z}$ and $q\in\mathcal{Q}$, $\beta(z)$ and $\beta(q)$ denote the values of each vertex $z$ and each edge $q$, respectively. Thus, the continuous state space of the deep network $\mathcal{M}$ can be defined by the set $\Omega=\left \{\beta(z),\beta(q): \forall z\in\mathcal{Z},q\in\mathcal{Q} \right \}$.
In this way, the total information contained in $\mathcal{M}$ can be measured by the entropy $E(\Omega)$ of set $\Omega$. 
The sets $E(\Omega_{z})=\left \{ \beta(z):z\in\mathcal{Z}\right \}$ and $E(\Omega_{q})=\left \{ \beta(q):q\in\mathcal{Q}\right \}$ represent the total information contained in the latent features and the network parameters, respectively. Specifically, $E(\Omega_{z})$ measures the feature representation power of the network $\mathcal{M}$ and $E(\Omega_{q})$ measures the complexity of the network. In our work, we focus on the entropy of latent features $E(\Omega_{z})$ rather than the entropy of network parameters $E(\Omega_{q})$. 

However, in the deep network $\mathcal{M}$, the latent features of the images always follow a complex and unknown distribution, so it is difficult to directly calculate the information entropy of latent features $E(\Omega_{z})$.
Fortunately, the Maximum Entropy Principle~\cite{cover1999elements} has proved that the entropy of a distribution is upper bounded by a Gaussian distribution with the same mean and variance,  as shown in Theorem 1. 

\begin{theorem}
For any continuous distribution $\mathbb{D}(a)$ of mean $\mu$ and variance $\sigma^{2}$, its differential entropy is maximized when $\mathbb{D}(a)$ is a Gaussian distribution $\mathcal{N}(\mu,\sigma^{2})$.
\end{theorem}

Therefore, 
we can estimate the upper bound of the entropy instead.
Suppose $a$ is sampled from a Gaussian distribution $\mathcal{N}(\mu,\sigma^{2})$, the differential entropy of $a$ can be defined as follows:
\begin{equation}
E(a)=\frac{1}{2}(\log(2\pi \sigma^{2})+1)=\frac{1}{2}(1+\log(2\pi)+\log(\sigma^{2})).
\label{Ea}
\end{equation}

As can be seen, the entropy of the Gaussian distribution depends only on the variance.
In this way, we can effectively approximate the entropy of the latent features $E(\Omega_{z})$ by simply computing the variance of the latent features. 

Different from previous works that adopt biased indicators to mine hard samples, we propose a novel hard sample mining strategy named Entropy-guided Hard Sample Mining (\textbf{EHSM}) to better achieve this goal.
EHSM comprehensively considers the local and global information of the tokens in measuring the difficulty of the sample.
Specifically, for an augmented sample $\widetilde{x}=(\widetilde{x}_{1},\widetilde{x}_{2},\cdots ,\widetilde{x}_{n})$, EHSM first estimates the local information entropy $E(\widetilde{x}_{i})=E(\widetilde{\kappa }_{i}\cdot \widetilde{f}_{i})$ of each local token $\widetilde{\kappa }_{i}\cdot \widetilde{f}_{i}$ using Eq.~\ref{Ea}. 
Then, all the local information entropy is aggregated as the global information entropy $\sum_{i}E(\widetilde{x}_{i})$ of the sample $\widetilde{x}$. Finally, EHSM utilizes an entropy-aware weight mechanism $\eta(\widetilde{x})=1+e^{-\gamma\sum_{i}E(\widetilde{x}_{i})}$ to adaptively assign importance weight to each sample, where $\gamma$ is a temperature coefficient.
The reweighted classification loss can be formulated as:
\begin{equation}
\mathcal{L}_{cls}^{trans}=\eta(\widetilde{x}) \times \mathcal{L}_{arc}(\widetilde{g},y).   
\end{equation}

It is worth mentioning that EHSM explicitly encourages the model to focus on
hard samples with less information.
In order to minimize the objective $\mathcal{L}_{cls}^{trans}$, the model has to optimize both the weight $\eta$ and the loss of basic classification $\mathcal{L}_{arc}$ during training, which will bring two benefits:
(1) Minimizing $\mathcal{L}_{arc}$ can encourage the model to learn better facial features from diverse training samples.
(2) Minimizing the weight $\eta(\widetilde{x})$ (\emph{i.e.}, maximizing the total information $\sum_{i}E(\widetilde{x}_{i})$) will facilitate the model to fully mine feature information contained in each face patch, especially some less-attended facial cues (\emph{e.g.}, nose, lip, and jaw), which significantly enhances the feature representation power of each local token. In this way, even if some important face patches are destroyed, the model can also make full use of the remaining facial cues to generalize the global token, leading to a more stable prediction.

\subsubsection{Optimization of TransFace Model}
To sum up, the overall objective of our TransFace can be formulated as follows:
\begin{equation}
\min \limits_{\Psi,\mathcal{F},\mathcal{S},\mathcal{C}}\mathcal{L}_{cls}^{trans}.
\end{equation}

\begin{figure*}[t]
\begin{center}
\includegraphics[width=0.8\linewidth]{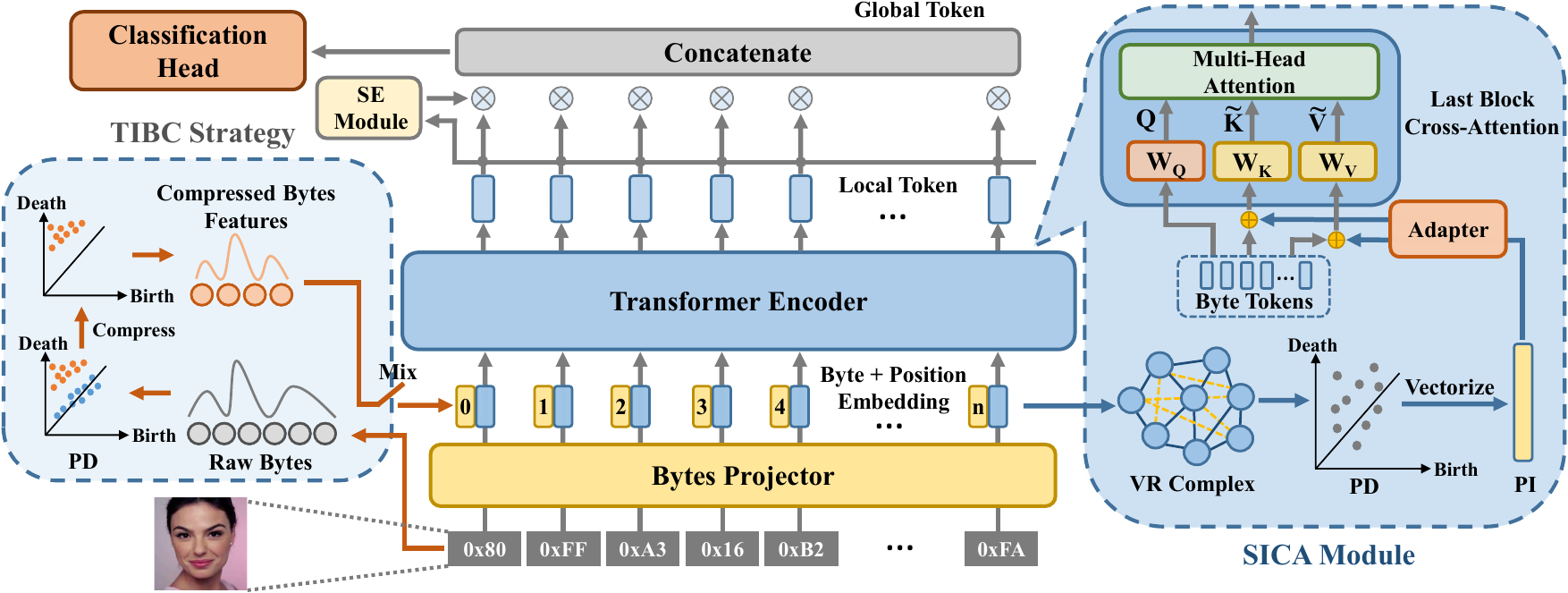}
\end{center}
\caption{Global overview of the proposed TransFace++ model. TIBC aims to extract compressed bytes features from raw bytes to mitigate information loss.
SICA establishes the interaction between byte tokens and structure information through a cross-attention operation. The adapter is utilized to adjust the dimensions and value range of PI representation.
$\oplus$ denotes the element-wise addition operation.
}
\label{fig-TransFace++}
\end{figure*}

\subsection{TransFace++}
\subsubsection{Preliminaries}
We utilize the TransFace architecture as the fundamental backbone network for our TransFace++ and make appropriate modifications to adapt it to handle image-byte inputs.
Specifically, as shown in Fig.~\ref{fig-TransFace++}, our TransFace++ model is composed of five components: a bytes projector $\mathcal{H}$, a transformer encoder $\mathcal{F}$, an adapter $\mathcal{J}$, an SE module $\mathcal{S}$, and a classification head $\mathcal{C}$. 
Mathematically, given an image $x$, we represent the image bytes as a sequence $\mathcal{B} = (b_{1}, b_{2}, \cdots, b_{N})$, where each $b_{i}$ denotes a byte and $N$ denotes the length of the image bytes. 
The bytes projector $\mathcal{H}$ is responsible for mapping bytes to embeddings $ \mathcal{E} = (h_{1}, h_{2}, \cdots, h_{n})$, where $n$ is the total number of byte embeddings.
The adapter $\mathcal{J}$ is employed to adjust the dimensions and value range of topological structure information in bytes.
Moreover, the output of the transformer encoder $\mathcal{F}$ is denoted as $(f_{1}, f_{2}, \cdots, f_{n})$. All local tokens extracted by $\mathcal{F}$ will pass through the SE module $\mathcal{S}$ and be rescaled as $( \kappa_{1}\cdot f_{1},\kappa_{2}\cdot f_{2},\cdots ,\kappa_{n}\cdot f_{n})$, where $\kappa_{1}, \cdots, \kappa_{n}$ represents the scaling factors generated by $\mathcal{S}$.
Then, all the rescaled local tokens are concatenated into a global token $z=\left [\kappa_{1}\cdot f_{1};\kappa_{2}\cdot f_{2}; \cdots \kappa_{n}\cdot f_{n} \right ]$, which will be fed into the classification head $\mathcal{C}$ for prediction. 

\subsubsection{Analysis}
Here, we first conduct several ablative experiments to investigate the intrinsic property of image bytes, which is presented in Table~\ref{tab-bytes-aug}.
From this comprehensive and fair analysis, several noteworthy observations come to light:
\textbf{(1)} ``Random Shuffle" encounters a sharp accuracy degradation as it severely disrupts the byte order.
\textbf{(2)} Compared with ``Random Shuffle", ``Strided Sampling" slightly boosts accuracy as it improves byte order consistency. 
\textbf{(3)} ``Window Shuffle" and ``Cyclic Shift" further enhance the accuracy as they preserve partial neighborhood relationships within the bytes.
\textbf{(4)} ``Reverse" effectively preserves the neighborhood relationships of bytes, leading to a comparable accuracy to the baseline model. These observations indicate that the ViT is highly sensitive to the \textbf{local correlation} of bytes rather than bytes frequencies (\emph{i.e.}, the number of different bytes). 

\begin{table}[t]
\caption{Verification accuracy (\%) of different bytes augmentation strategies on IJB-C benchmark~\cite{maze2018iarpa}. Training Data: MS1MV2. \textbf{(1) Random Shuffle:} Randomly shuffling the byte order during training. \textbf{(2) Strided Sampling:} A strided sampling with a stride size of 512 (\emph{e.g.}, [0, 512, 1024, $\cdots$, 1, 513, 1025, $\cdots$]) is performed. \textbf{(3) Window Shuffle:} Randomly swapping the order of different windows, each containing 512 bytes. \textbf{(4) Cyclic Shift:} The latter half of the bytes are shifted to the beginning. \textbf{(5) Reverse:} The byte order is reversed. 
Each strategy will be applied to the image bytes with a probability of 60\% during training.
} 
\centering
\resizebox{1\linewidth}{!}{
\begin{tabular}{c|c| c c c }
\hline
Method & Data Format & Augmentation & IJB-C(1e-5) & IJB-C(1e-4) \\
\hline \hline 
 &  & Random Shuffle & 53.41 & 74.25 \\
 &  & Strided Sampling & 62.74 & 79.56 \\
ViT-S & TIFF & Window Shuffle & 75.16 & 83.30 \\
 &  & Cyclic Shift & 85.89  & 92.58 \\
 &  & Reverse & 87.38 & 93.76 \\
\cline{1-5}
ViT-S (Baseline) & TIFF & - & \textbf{91.49} & \textbf{94.74} \\
\hline 
\end{tabular}
}
\label{tab-bytes-aug}
\end{table}

\subsubsection{Topology-based Image Bytes Compression}

Compared to using RGB pixels as input, the main challenge of using image bytes as input is the excessively long sequence length. Table~\ref{tab-byte-length} shows the length of the bytes of different encodings. We can observe that the input shape $\mathbb{S}$ of different image byte encodings exceeds 37,600 tokens, which poses significant difficulties for the self-attention module in ViT due to its quadratic computational cost.
Most existing work usually adopts window-shifted attention strategies~\cite{liu2021swin, hortonbytes,fang2022msg} to alleviate the computational burden of the model.
However, these strategies inevitably destroy the local correlation of bytes and sever the connections between them, which seriously affects the model's understanding of image bytes and causes the FR model to optimize in the incomplete direction. Hence, we still utilize the global attention operation in our TransFace++ to fully preserve the local correlation of bytes. 

\begin{figure}[t]
\begin{center}
\includegraphics[width=1\linewidth]{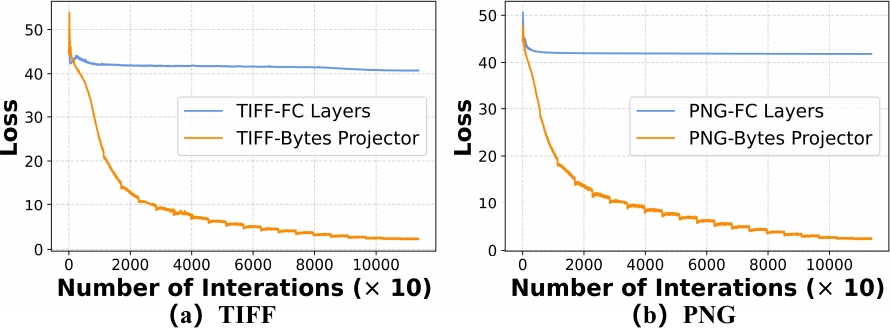}
\end{center}
   \caption{Comparison of training loss using bytes projector $\mathcal{H}$ and FC layers on TIFF and PNG encodings.
   }
\label{fig2}
\end{figure}

\noindent
{\textbf{Bytes Projector.}}
To alleviate the substantial computational burden imposed by long sequential data, we devise a byte{s} projector $\mathcal{H}$ that consists of a learnable embedding layer (vocabulary size = $2^{8}$) and two Conv1D layers.
Specifically, the learnable embedding layer is employed to map input bytes into byte embeddings $\mathcal{E}$, while the Conv1D layers are utilized to reduce the sequence length (from 37,600+ to 144).
In preliminary experiments, we attempted to simply utilize fully connected (FC) layers to map image bytes and reduce the sequence length. However, this approach results in convergence difficulties for the model during training, as shown in Fig.~\ref{fig2}.
This is because FC layers struggle to preserve the local correlation within image bytes during the mapping process, while Conv1D layers excel at capturing and preserving these neighborhood relationships.
{Notably, the experimental results in Fig.~\ref{fig2} further demonstrate that the Conv1D-based bytes mapping mechanism (reducing dimensionality from 37,600+ to 144) effectively preserves key facial features, thus ensuring that the model converges reliably and stably.
}

Although directly mapping image bytes to short-length byte embeddings effectively alleviates the model's overall computational load and improves memory usage, it inevitably results in substantial information loss, which may seriously affect the learning of discriminative facial features.
To remedy this issue, motivated by signal compression theory~\cite{gersho2012vector,edelsbrunner2006persistence}, we introduce a Topology-based Image Bytes Compression (\textbf{TIBC}) strategy to minimize information loss. 
Mathematically, a one-dimensional signal is typically represented as a function with real value $\mathcal{W}$ defined in a closed time interval $\left [ T_{1}, T_{n}\right ]$. In this work, we model image bytes $\mathcal{B} = (b_{1}, b_{2}, \cdots, b_{n})$ as a one-dimensional time series signal at a set of discrete sampling time points $T_{1}=t_{1}< t_{2}< t_{3}< \cdots < t_{n}=T_{n}$. Here, the value of byte $b_{i}$ represents the signal intensity at the time point $t_{i}$.
As illustrated in Fig.~\ref{fig-TransFace++}, our TIBC strategy mainly comprises two steps: (1) computing the signal's Persistence Diagram (PD), and (2) topologically compressing the signal.

\begin{table}
\caption{Bytes length of different image byte encodings.
$\mathbb{S}$ represents the input shape, and $n$ denotes the token length that is passed to the transformer encoder.
}
\centering
\resizebox{0.85\linewidth}{!}{
\begin{tabular}{c|c c c }
\hline
Method & Data Format & $\mathbb{S}$ & $n$ \\
\hline \hline 
ViT / \textbf{TransFace++} & fHWC & 37632 & 144  \\
ViT / \textbf{TransFace++} & fCHW & 37632 & 144  \\
ViT / \textbf{TransFace++} & TIFF & 37772 & 144  \\
ViT / \textbf{TransFace++} & PNG & 37817 & 144  \\
\cline{1-4} %
ViT & RGB & $3 \times 112 \times 112$ & 144  \\
\hline 
\end{tabular}
}
\label{tab-byte-length}
\end{table}

\noindent
\textbf{(1) Signal's PD.}
Previous studies~\cite{edelsbrunner2002topological,chazal2009proximity} have demonstrated that the one-dimensional signal $\mathcal{B}$ can be summarized by its 0-dimensional PD, \emph{i.e.}, $\mathrm{D}_{0}(\mathcal{B})$, which is obtained by vertically scanning the horizontal line from negative infinity to positive infinity to track connected components. 
The resulting PD $\mathrm{D}_{0}(\mathcal{B})$ encodes the lifespan information $(\varpi, \tau)$ of the components, where the components emerge at local minima and disappear at local maxima. 
Specially, low-persistence components usually are noncritical points, and their persistence points $(\varpi, \tau)$ are close to the diagonal of PD~\cite{edelsbrunner2002topological}.
As a result, by tracking the birth time $\varpi$ and death time $\tau$ of the components, we can separately extract the critical points (\emph{i.e.}, prominent features) and the non-critical points (\emph{i.e.}, noise or redundant information) in the signal. 

\noindent
\textbf{(2) Topological Compression.}
The classical Morse Cancellation Lemma~\cite{milnor1963morse} has demonstrated that if $(\varpi, \tau)$ is the lowest-persistence point in
the PD $\mathrm{D}_{0}(\mathcal{W})$ of signal $\mathcal{W}$, there exists a signal $\mathcal{K}$ defined on the same domain with a PD $\mathrm{D}_{0}(\mathcal{K})$ that is identical to $\mathrm{D}_{0}(\mathcal{W})$ except for the removal of $(\varpi, \tau)$, which means that signal $\mathcal{K}$ can be created by ``un-kinking" the low-persistence points in $\mathcal{W}$.
Based on this lemma and considering the fact that low persistence points correspond to noise or redundant information in the signal, we can generate the compressed signal $\mathcal{B}^{\delta}$ (\emph{i.e.}, compressed bytes features) by discarding the low-persistence points (\emph{i.e.}, non-critical points). In this way, the compressed byte feature $\mathcal{B}^{\delta}$ is a subset of the raw bytes $\mathcal{B}$, which simply discards some noise or redundant information while retaining valuable prominent features of the original image bytes (see Fig.~\ref{fig3} for visualization). 
The process can be formulated as:
\begin{equation}
\mathcal{B}^{\delta} =\mathrm{TIBC}(\mathcal{B}). 
\end{equation}

Next, to mitigate information loss during the mapping process, we integrate the compressed byte feature $\mathcal{B}^{\delta}$ with byte embeddings $\mathcal{E}$ in the following manner:
\begin{equation}
g = \mathcal{E} + \mathcal{B}^{\delta}.
\end{equation}

{
The motivation for using addition to reintegrate features can be found in Sec.~3.2 of the Supplementary Materials.}
Then, we feed the mixed embeddings $g$ into the transformer encoder for supervised training.
In particular, unlike conventional Fourier-based signal compression methods~\cite{karim2011wavelet,gilbert2014recent} that primarily focus on reconstructing an approximate version of the original signal, our TIBC retains some crucial information from the original signal in the compressed features, effectively preserving important facial cues.

\noindent
\textbf{Probabilistic Mask Trick.}
As training progresses, we unexpectedly find that the model increasingly relies on the TIBC branch $\mathcal{B}^{\delta}$, severely hampering the optimization of the bytes projector $\mathcal{H}$.
Inspired by Dropout, we introduce a probabilistic mask $p_{\mathcal{M}}$ to randomly mix compressed features and byte embeddings during training:
\begin{equation}
g = \mathcal{E} + p_{\mathcal{M}}(\mathcal{B}^{\delta}),
\end{equation}
where
\begin{equation}
p_{\mathcal{M}}(a)=
\begin{cases}
0,  & \lambda_{p} > \mu_{p} \\
a,    & \lambda_{p} \le \mu_{p} 
\end{cases}
\end{equation}
$\lambda_{p} \sim \mathcal{U}(0, 1)$, and $\mathcal{U}(0, 1)$ is a uniform distribution on $\left [0, 1\right]$. We choose $\mu_{p}=0.3$.
This trick not only effectively directs the model's focus towards optimizing the bytes projector branch but also reduces the risk of the model overfitting to the input bytes. 

\noindent
\textbf{Classification Loss.}
In our TransFace++ model, we also adopt the widely used ArcFace~\cite{deng2019arcface} as the classification loss function:
\begin{equation}
    \mathcal{L}_{cls}^{byte}(\mathcal{B},y)=-\log\frac{e^{s(\cos(\theta_{y}+m))}}{e^{s(\cos(\theta_{y}+m))}+\sum_{c=1,c\neq y}^{C}e^{s\cos\theta_{c}}}
\end{equation}
where $y$ represents the class label of the input image bytes $\mathcal{B}$, $s$ is a scaling hyperparameter, $\theta_{y}$ indicates the angel between the center of the $c$-th class and the feature, $m>$0 is an additive angular margin and $C$ denotes the class number.

\subsubsection{Structure Information-guided Cross-Attention}
As mentioned in Sec.~\ref{intro}, 
face image bytes are inherently sequential data and lack the geometric information presented in the RGB images, which severely hinders the model’s understanding of the image bytes.
To address this problem, an intuitive idea is to supplement the missing structure information for the image bytes.
It is worth mentioning that several studies~\cite{carlsson2009topology,moor2020topological} on topology have demonstrated that regardless of the space in which the data exist, its intrinsic data structure should remain unchanged.
As a result, for a given image, both the RGB pixels format data and the image bytes data should contain the same topological structure information.

Inspired by the invariant nature of data structure, we propose a novel Structure Information-guided Cross-Attention (\textbf{SICA}) module to establish the interaction between byte tokens and structure information.
Specially, SICA first leverages Persistent Homology (PH) to mine the inherent topological structure information in image bytes and then injects it into byte tokens through a cross-attention operation. 

For the image bytes $\mathcal{B} = (b_{1}, b_{2}, \cdots, b_{n})$, the byte embeddings $\mathcal{E} = (h_{1}, h_{2}, \cdots, h_{n})$ extracted by the projector $\mathcal{H}$ comprise $n$ high-dimensional data points and thus can be viewed as a point cloud in a metric space.
Based on the pairwise distance $\phi(h_{i},h_{j})$ (\emph{e.g.}, Euclidean distance) between points,
we can construct a VR complex $\mathcal{VR}_{\epsilon}(\mathcal{E})$ for the point cloud $\mathcal{E}$.
After that, we can compute the PH of complex $\mathcal{VR}_{\epsilon}(\mathcal{E})$, and obtain its corresponding PD $\mathrm{D}_{0}(\mathcal{E})$ which encodes rich topological structure information about image bytes $\mathcal{B}$.
However, since PD is a multiset of points rather than a linear vector, it is not suitable for being used directly as input for general machine learning pipelines~\cite{som2018perturbation}.
Therefore, we convert the PD into a finite-dimensional vector representation (\emph{i.e.}, PI) $\mathrm{I}$~\cite{adams2017persistence}. This vector representation $\mathrm{I}$ provides a stable representation of the topological characteristics captured by the original PD. In this way, we can effectively mine the underlying structure information in image bytes.

Next, we shift our focus towards injecting structure information into byte tokens to strengthen the model’s understanding of image bytes. The self-attention module is a critical component in Transformer~\cite{dosovitskiy2020image}, which aims to highlight the correlations between all entities in a sequence.
For self-attention, the byte tokens $\mathrm{T}$ are first projected into three vectors, \emph{i.e.}, queries $\mathrm{Q}$, keys $\mathrm{K}$, and values $\mathrm{V}$:
\begin{equation}
\mathrm{Q} = \mathrm{T}\mathrm{W}_\mathrm{Q}, \mathrm{K} =\mathrm{T}\mathrm{W}_\mathrm{K}, \mathrm{V} =\mathrm{T}\mathrm{W}_\mathrm{V},
\end{equation}
where $\mathrm{W}_\mathrm{Q}$, $\mathrm{W}_\mathrm{K}$, and $\mathrm{W}_\mathrm{V}$ are learnable parameters.  
Then the output of self-attention is obtained by calculating a weighted sum of the values using the following formula:
\begin{equation}
Attn_{self}(\mathrm{Q},\mathrm{K},\mathrm{V})=softmax(\frac{\mathrm{Q}\mathrm{K}^{\top}}{\sqrt{r_{k}}})\mathrm{V}. 
\end{equation}

The cross-attention module is a variant of the self-attention module which aims to achieve the interaction of knowledge from different modalities.
Considering that image bytes and their structure information are heterogeneous data, it is essential to map these data into the unified space before establishing an effective interaction.
To achieve this goal, the PI representation $\mathrm{I}$ is first fed into the adapter $\mathcal{J}$ to adjust its dimensions and range of values:
\begin{equation}
\mathrm{I}_{A}=\mathcal{J}(\mathrm{I}).
\end{equation}
Then it is injected into the keys $\widetilde{\mathrm{K}}$ and the values $\widetilde{\mathrm{V}}$ of the byte tokens $\mathrm{T}$ to establish the interaction between the byte tokens and the structure information. 
The detailed interaction process of the cross-attention module in our TransFace++ can be defined as follows:
\begin{equation}
\widetilde{\mathrm{K}} =(\mathrm{T}\oplus  \mathrm{I}_{A})  \mathrm{W}_\mathrm{K}, \widetilde{\mathrm{V}} 
=(\mathrm{T}\oplus  \mathrm{I}_{A})\mathrm{W}_\mathrm{V}, 
\end{equation}
\begin{equation}
Attn_{cross}(\mathrm{Q},\widetilde{\mathrm{K}},\widetilde{\mathrm{V}})=softmax(\frac{\mathrm{Q}(\widetilde{\mathrm{K}})^{\top}}{\sqrt{r_{k}}}) \widetilde{\mathrm{V}},
\end{equation}
where $\oplus$ denotes the element-wise addition operation.
In this way, topological structure information is successfully injected into the image bytes,
greatly enhancing the model's understanding of bytes information.
In practice, we only apply the SICA module in the last block of the transformer encoder to achieve information interaction ({see Sec.~3.2 of the Supplementary
Material for a detailed analysis}).

\subsubsection{Optimization of TransFace++ Model}
In summary, the overall objective of TransFace++ can be expressed as:
\begin{equation}
\min \limits_{\mathcal{H},\mathcal{F},\mathcal{C},\mathcal{J}}\mathcal{L}_{cls}^{byte}.
\end{equation}

\begin{table*}[!t]
  \caption{Verification accuracy (\%) on LFW, CFP-FP, AgeDB-30, IJB-C and IJB-B benchmarks. 
  $^{\dagger}$ denotes the conference version \cite{dan2023transface} of our model.
  }
  \label{table-transface}
  \centering
  \resizebox{0.75\linewidth}{!}{
  \begin{tabular}{c|c|c|ccc|cc|c}
    \hline
    \multirow{2}{*}{Training Data} & 
    \multirow{2}{*}{Method} &
    \multirow{2}{*}{Data Format} &
    \multirow{2}{*}{LFW} &
    \multirow{2}{*}{CFP-FP} &
    \multirow{2}{*}{AgeDB-30} &
    \multicolumn{2}{c}{IJB-C} &
    \multicolumn{1}{|c}{IJB-B} 
    \\\cline{7-9}
    & & & & &  & 1e-5 & 1e-4 & 1e-4  \\
    \hline \hline
    & R50, ArcFace \cite{deng2019arcface}  &  & 99.68 & 97.11 & 97.53  & 88.36 & 92.52 & 91.66 \\

    & R50, MagFace \cite{meng2021magface}  &  & 99.74 & 97.47 & 97.70   & 88.95 & 93.34 & 91.47 \\

    & {R50, AdaFace} \cite{kim2022adaface}  & RGB & {99.82} & {97.86} & {97.85}   & {-} & {96.27} & {94.42} \\

    & {R50, TopoFR \cite{dan2024topofr}}  &  & \textbf{{99.83}} & \textbf{{98.24}} & \textbf{{98.23}}  & \textbf{{94.79}} & {96.42} & {95.13} \\

    & ViT-S  &  & 99.62 & 97.18 & 97.45 & 93.40 & 95.89  & 92.63  \\
    & $^{\dagger}$\textbf{TransFace-S}  &  & 99.82 & 98.04 & 97.98 & 93.87 & \textbf{96.45}  & \textbf{95.44}  \\

    \cline{2-9} 
    & R100, CosFace \cite{wang2018cosface} &  & 99.78 & 98.26 & 98.17   & 92.68 & 95.56 & 94.01 \\
    & R100, ArcFace \cite{deng2019arcface} &  & 99.77 & 98.27 & 98.15   & 92.69 & 95.74  & 94.09 \\
    & R100, MV-Arc-Softmax \cite{wang2020mis} &  & 99.80 & 98.28 & 97.95  &- & 95.20 & 93.60 \\
    & R100, BroadFace \cite{kim2020broadface} &  & \textbf{99.85} & 98.63 & 98.38   & 94.59 & 96.38 & 94.97 \\
     & R100, CurricularFace \cite{huang2020curricularface} &  & 99.80 & 98.37 & 98.32   &- & 96.10 & 94.80 \\
    & R100, MagFace+ \cite{meng2021magface} &  &  99.83 & 98.46 & 98.17  & 94.08 & 95.97  & 94.51 \\
    MS1MV2 & R100, SCF-ArcFace \cite{li2021spherical} & RGB & 99.82 & 98.40 & 98.30  & 94.04 & 96.09 & 94.74 \\
    & R100, DAM-CurricularFace \cite{liu2021dam} &  & - & -  &- &- & 96.20 & 95.12  \\

    & R100, ElasticFace-Cos+ \cite{Boutros_2022_CVPR} &  & 99.80 & \textbf{98.73} & 98.28  &- & 96.65 & 95.43 \\

    & {R100, AdaFace} \cite{kim2022adaface} &  & {99.82} & {98.49} & {98.05}  & {-} & {96.89} & {95.67} \\

    & {R100, TopoFR} \cite{dan2024topofr}  &  & \textbf{{99.85}} & {98.71} & \textbf{{98.42}} & \textbf{{95.23}} & \textbf{{96.95}}  & {95.70} \\

    & ViT-B  &  & 99.76 & 98.25 & 98.02 & 94.08 & 96.15  & 94.92 \\

    & $^{\dagger}$\textbf{TransFace-B}  &  & 99.83 & 98.66 & 98.37  & 94.15 & 96.55 & 95.63 \\

    & \textbf{TransFace-B}  &  & \textbf{99.85} & 98.70 & 98.40  & 94.66 & 96.78 & \textbf{95.79} \\

    
    \cline{2-9}
    & R200, ArcFace \cite{deng2019arcface} &  & 99.80 & 98.38 & 98.18  & 94.24 & 96.31 &  95.18 \\
    & R200, MagFace+ \cite{meng2021magface} &   &99.83 & 98.62 & 98.26  & 94.38 & 96.43 & 95.35 \\

    & {R200, AdaFace} \cite{kim2022adaface}  &  & {99.83} & {98.76} & {98.28}   & {94.88} & {96.93} & {95.71} \\

    & {R200, TopoFR} \cite{dan2024topofr} & RGB & \textbf{{99.85}} & \textbf{{99.05}} & \textbf{{98.52}}  & \textbf{{95.15}} & \textbf{{97.08}} & {95.82} \\

    & ViT-L  &  & 99.79 & 98.42 & 98.17 & 94.11 & 96.24  & 95.16  \\
    
    & $^{\dagger}$\textbf{TransFace-L} &   & \textbf{99.85} & 98.76  & 98.43  & 94.55 & 96.59  & 95.74 \\

    & \textbf{TransFace-L} &   & \textbf{99.85} & 98.83  & 98.46  & 94.97 & 96.88  & \textbf{95.92} \\
    
    \hline \hline
    & R50, ArcFace \cite{deng2019arcface} &  & 99.78 & 98.77 & 98.28  & 95.29 & 96.81 & 95.30  \\
    & R50, MagFace+ \cite{meng2021magface} &   & 99.80 & 98.82 & 98.33  & 95.56 & 96.90  & 95.61  \\

    & {R50, AdaFace} \cite{kim2022adaface} &   & {99.82} & {99.07} & {98.34}  & {95.58} & {96.90}  & {95.66}  \\

    & {R50, TopoFR} \cite{dan2024topofr} & RGB  & \textbf{{99.85}} & \textbf{{99.28}} & {98.47}  & {95.99} & {97.27}  & {95.96}  \\

    & ViT-S  &  & 99.80 & 98.85 & 98.24 & 95.24 & 96.70  & 95.52  \\

    & {ViT-S, LVFace} \cite{you2025lvface}  &  & {-} & {-} & {-} & \textbf{{96.52}} & {97.31}  & \textbf{{96.14}}  \\
    
    & $^{\dagger}$\textbf{TransFace-S} &   & \textbf{99.85} & 98.91 & 98.50  & 96.06 & 97.33  & 96.05  \\

    & \textbf{TransFace-S} &   & \textbf{99.85} & 98.96 & \textbf{98.56}  & 96.10 & \textbf{97.45}  & 96.12  \\

    \cline{2-9}
     & R100, ArcFace \cite{deng2019arcface} &  & 99.81 & 99.04 & 98.31  & 95.38 & 96.89 & 95.69 \\
    & R100, MagFace+ \cite{meng2021magface} &  & 99.83 & 99.08 & 98.40  & 95.73 & 97.04 & 95.84 \\

    & {R100, AdaFace} \cite{kim2022adaface} &   & {99.82} & {99.20} & {98.58}  & {96.24} & {97.19}  & {95.87}  \\

    & {R100, TopoFR} \cite{dan2024topofr} & RGB  & \textbf{{99.85}} & \textbf{{99.43}} & \textbf{{98.72}}  & {96.57} & {97.60}  & {96.34}  \\

    Glint360K & ViT-B  &  & 99.82 & 99.02 & 98.33 & 95.41 & 96.88  & 95.70  \\

    & {ViT-B, LVFace} \cite{you2025lvface} &  & {-} & {-} & {-}  & \textbf{{97.00}} & \textbf{{97.70}} & \textbf{{96.51}} \\

    & $^{\dagger}$\textbf{TransFace-B} &  & \textbf{99.85} & 99.17 & 98.53  & 96.18 & 97.45 & 96.33 \\

    & \textbf{TransFace-B} &  & \textbf{99.85} & 99.24 & 98.62  & 96.25 & 97.59 & 96.46 \\

    \cline{2-9}
    & R200, ArcFace \cite{deng2019arcface} &  & 99.82 & 99.14 & 98.49  & 95.71 & 97.20 & 95.89 \\
    & R200, MagFace+ \cite{meng2021magface} &   &99.83 & 99.20 & 98.55  & 95.96 & 97.33 & 96.12  \\

    & {R200, AdaFace} \cite{kim2022adaface} &   & {99.83} & {99.24} & {98.61}  & {95.96} & {97.33}  & {96.12}  \\

    & {R200, TopoFR} \cite{dan2024topofr} &  RGB  & \textbf{{99.87}} & \textbf{{99.45}} & \textbf{{98.82}}  & {96.71} & \textbf{{97.84}}  & {96.56}  \\

    & ViT-L  &  & 99.82 & 99.10 & 98.47 & 95.78 & 97.13  & 95.85  \\

    & {ViT-L, LVFace} \cite{you2025lvface} &  & {-} & {-} & {-}  & \textbf{{97.02}} & {97.66} & {96.51} \\
    
    & $^{\dagger}$\textbf{TransFace-L}  &  & 99.85 & 99.32 & 98.62   & 96.29 & 97.61 & 96.64 \\

    & \textbf{TransFace-L}  &   & 99.85 & 99.37 & 98.66   & 96.33 & 97.80 & \textbf{96.73} \\

    \hline \hline
    & R50, ArcFace \cite{deng2019arcface} &  & 99.80 & 99.04 & 98.35  & 95.86 & 97.31 & 95.94  \\
    
    & R50, MagFace+ \cite{meng2021magface} & RGB  & 99.83 & 99.08 & 98.47  & 96.13 & 97.48  & 96.17   \\

    & ViT-S  &   & 99.83 & 99.15 & 98.40 & 95.92 & 97.30  & 95.91  \\
    
    & \textbf{TransFace-S} &  & \textbf{99.85} & \textbf{99.30} & \textbf{98.65}  & \textbf{96.66} & \textbf{97.85}  & \textbf{96.52}  \\

    \cline{2-9}
     & R100, ArcFace \cite{deng2019arcface} &  & 99.82 & 99.19 & 98.45  & 96.01 & 97.40 & 96.06 \\
    & R100, MagFace+ \cite{meng2021magface} & RGB & 99.83 & 99.24 & 98.56  & 96.30 & 97.61 & 96.32 \\

    & ViT-B  &  & \textbf{99.85} & 99.22 & 98.43  & 96.07 & 97.48  & 96.10  \\
    
    WebFace42M & \textbf{TransFace-B} &  & \textbf{99.85} & \textbf{99.45} & \textbf{98.69}  & \textbf{97.25} & \textbf{97.96} & \textbf{96.67} \\

    \cline{2-9}
    & R200, ArcFace \cite{deng2019arcface} &  & 99.83 & 99.25 & 98.48  & 96.35 & 97.61 & 96.18 \\
    
    & R200, MagFace+ \cite{meng2021magface} &   & \textbf{99.85} & 99.36 & 98.60  & 96.67 & 97.79 & 96.43  \\

    & R200, UniFace \cite{zhou2023uniface} &  & 99.83 & 99.42 & 98.66  & 96.88 & 97.98 & -  \\

    & {R200, UniTSFace} \cite{li2023unitsface}  & RGB & {99.83} & {99.47} & {98.71} & {97.00} & {97.99}  & {-}  \\

    & {R200, TopoFR} \cite{dan2024topofr}  &  & {99.80} & {99.31} & {98.70} & {97.10} & {98.01}  & {-}  \\
    
    & ViT-L  &  & 99.83 & 99.25 & 98.36 & 96.46 & 97.65  & 96.27  \\

    & {ViT-L, LVFace} \cite{you2025lvface}  &  & {99.80} & {99.41} & {98.66} & {97.25} & {98.06}  & {-}  \\

    & \textbf{TransFace-L}  &   & \textbf{99.85} & \textbf{99.50} & \textbf{98.72}   & \textbf{97.37} & \textbf{98.16} & \textbf{96.80} \\
    \hline
  \end{tabular} }
\end{table*}

\section{Experiments}
\subsection{Setup} 
\noindent
\textbf{Training Datasets.} 
We separately adopt MS1MV2~\cite{deng2019arcface} dataset (5.8M images, 85K identities), Glint360K~\cite{an2021partial} dataset (17M images, 360K identities), the recently proposed {extremely} large-scale WebFace42M~\cite{zhu2021webface260m} dataset (42.5M images, 2M identities), {and the synthetic SFace-60 dataset (634K images, 10.5K identities)~\cite{boutros2022sface}} to train our model. 

\noindent
\textbf{Test Datasets.}
For evaluation, we employ LFW~\cite{huang2008labeled}, AgeDB-30~\cite{moschoglou2017agedb}, CFP-FP~\cite{sengupta2016frontal}, IJB-C~\cite{maze2018iarpa}, IJB-B~\cite{whitelam2017iarpa}, CPLFW~\cite{zheng2018cross}, CALFW~\cite{zheng2017cross}, TinyFace~\cite{cheng2019low}, MegaFace~\cite{kemelmacher2016megaface}, {RFW~\cite{wang2019racial}, {and the ICCV-2021 Masked Face Recognition Challenge (\textbf{MFR-Ongoing})~\cite{deng2021masked}} as benchmarks to test model performance.}

\noindent
Note that we denote the conference version~\cite{dan2023transface}
of our TransFace models by $\dagger$ in Table~\ref{table-transface}.
To unlock the potential of our TransFace models, we employ grid search to find the most suitable hyper-parameters (\emph{e.g.}, the number of selected dominant patches $K_{0}$, learning rate, batch size, the number of GPUs, and training epochs) for different architectures and different training sets.
By fine-tuning these hyper-parameters, we can facilitate better convergence of these models and further enhance their generalization performance.
Due to page size limitation, more training settings and implementation details are placed in the Supplementary Materials.

\begin{table*}[!htbp]
  \caption{Verification accuracy (\%) on LFW, CFP-FP, AgeDB-30, IJB-C and IJB-B benchmarks.
  {Note that ViT baseline simply introduces the bytes projector $\mathcal{H}$ to process image byte inputs, mapping input byte sequences longer than 37,600 into compact embeddings of length 144.}
  }
  \label{table-TransFace++}
  \centering
  \resizebox{0.799\linewidth}{!}{
  \begin{tabular}{c|c|c|ccc|cc|c}
    \hline
    \multirow{2}{*}{Training Data} & 
    \multirow{2}{*}{Method} &
    \multirow{2}{*}{Data Format} &
    \multirow{2}{*}{LFW} &
    \multirow{2}{*}{CFP-FP} &
    \multirow{2}{*}{AgeDB-30} &
    \multicolumn{2}{c}{IJB-C} &
    \multicolumn{1}{|c}{IJB-B} 
    \\\cline{7-9}
    & & & & &  & 1e-5 & 1e-4 & 1e-4  \\
    \hline \hline

    & R100, ArcFace \cite{deng2019arcface} &  & 88.43 & 86.61 & 84.03 & 31.07 & 60.23 & 58.29 \\
    & ViT-S &  & 99.46  & 96.60 & 97.11 & 91.72 & 94.46 & 92.83 \\
    & ViT-B &  & 99.61 & 96.78 & 97.27 & 91.90 & 94.77 & 93.25 \\
     & ViT-L & fHWC & 99.70 & 97.03 & 97.42 & 92.34 & 95.02 & 93.64 \\
     & \textbf{TransFace++-S} &  & \textbf{99.85} & 98.01 & 97.96 & 93.81 & 96.25 & 95.18 \\
    & \textbf{TransFace++-B} &  & \textbf{99.85} & 98.61 & 98.36 & 94.36 & 96.53 & 95.40 \\
    & \textbf{TransFace++-L} &  & \textbf{99.85} & \textbf{98.72} & \textbf{98.44}  & \textbf{94.69} & \textbf{96.77} & \textbf{95.62} \\
    
     \cline{2-9}
     & R100, ArcFace \cite{deng2019arcface} &  &  87.71 & 84.39  & 82.14 & 29.37 & 57.59 & 54.83 \\
    & ViT-S &  & 99.33 & 95.47 & 96.85 & 91.31 & 94.22 & 92.57 \\
    & ViT-B &  & 99.49 & 95.76 & 97.03 & 91.68 & 94.50 & 92.88 \\
    & ViT-L & fCHW & 99.62 & 96.14 & 97.25  & 92.07 & 94.89 & 93.34  \\
    & \textbf{TransFace++-S} &  & 99.83 & 97.68 & 97.81 & 93.55 & 96.13 & 95.10 \\
    & \textbf{TransFace++-B} &  & 99.84 & 98.37 & 98.15 & 94.14 & 96.36 & 95.24 \\
    & \textbf{TransFace++-L} &  & \textbf{99.85} & \textbf{98.54} & \textbf{98.32} & \textbf{94.57} & \textbf{96.60} & \textbf{95.53} \\
    
    \cline{2-9}
    MS1MV2 & R100, ArcFace \cite{deng2019arcface} &  & 88.19 & 85.52 & 82.79 & 30.03 & 58.23 & 57.05 \\
     & ViT-S &  & 99.52 & 96.70 & 97.22 & 91.49 & 94.74 & 93.20 \\
    
    & ViT-B &  & 99.66 & 96.81 & 97.38 & 91.93 & 94.96 & 93.47 \\
    
     & ViT-L & TIFF & 99.76 & 97.10 & 97.62 & 92.48 & 95.25 & 93.74 \\
    & \textbf{TransFace++-S} &  & \textbf{99.85} & 98.05 & 97.98 & 94.10 & 96.33 & 95.21 \\
    & \textbf{TransFace++-B} & & \textbf{99.85} & 98.53 & 98.34 & 94.38 & 96.67 & 95.45 \\
    & \textbf{TransFace++-L} &  & \textbf{99.85} & \textbf{98.72} & \textbf{98.40} & \textbf{94.79} & \textbf{96.85} & \textbf{95.66} \\

     \cline{2-9}
    & R100, ArcFace \cite{deng2019arcface} &  & 86.55 & 82.48 & 80.63 & 27.43 & 56.32 & 52.71 \\
    & ViT-S &  & 99.27 & 93.55 & 95.06 & 87.62 & 92.88 & 91.85 \\
    & ViT-B &  & 99.45 & 93.80 & 95.27 & 87.96 & 93.24 & 91.36 \\
     & ViT-L & PNG & 99.56 & 94.52 & 95.71 & 88.68 & 93.75 & 92.15 \\
    
    & \textbf{TransFace++-S} &  & 99.82 & 96.83 & 97.78 & 92.30 & 95.45 & 94.02 \\
    & \textbf{TransFace++-B} &  & \textbf{99.83} & 97.32 & 98.05 & 92.61 & 95.84 & 94.35 \\
    
    & \textbf{TransFace++-L} &  & \textbf{99.83} & \textbf{97.76} & \textbf{98.21} & \textbf{94.14} & \textbf{96.37} & \textbf{95.26} \\
    \hline
  \end{tabular} }
\end{table*}

\subsection{Results on Mainstream Benchmarks}

\subsubsection{Results of TransFace}

\noindent
\textbf{Results on LFW, CFP-FP, and AgeDB-30.}
We train our TransFace on MS1MV2, Glint360K and WebFace42M, respectively, and compare it with state-of-the-art {(SOTA)} competitors on three benchmarks, as reported in Table~\ref{table-transface}.
It is worth mentioning that the performance of the existing FR models on these three benchmarks has reached saturation. 
We find that the ViT baseline model does not perform well in FR tasks, and its performance is almost on par with the ResNet-based baseline model. 
{
While our TransFace achieves a clear increase in accuracy compared to the ViT baseline and becomes cutting-edge models across different backbone architectures.
}

{
Specifically, on the MS1MV2 and Glint360K training sets, TransFace achieves accuracy comparable to ResNet-based SOTA competitors TopoFR~\cite{dan2024topofr} and AdaFace~\cite{kim2022adaface}, and even surpasses them on certain metrics, demonstrating superior generalization ability. 
More importantly, when trained on the large-scale WebFace dataset, TransFace fully unlocks its powerful representation potential, outperforming both the ResNet-based SOTA method TopoFR~\cite{dan2024topofr} and the ViT-based SOTA method LVFace~\cite{you2025lvface}, thereby establishing itself as a leading framework.
}

\noindent
\textbf{Results on IJB-C and IJB-B.}
{As shown in Table~\ref{table-transface}, our three TransFace models trained with the MS1MV2 dataset significantly outperform most advanced competitors on IJB-B and IJB-C benchmarks in terms of accuracy, and approach the level of the ResNet-based SOTA method TopoFR~\cite{dan2024topofr} and the ViT-based SOTA method LVFace~\cite{you2025lvface}. 
For instance, on the MS1MV2 training set, TransFace-S achieves improvements of +0.03\% on IJB-C (TAR@FAR=1e-4) and +0.31\% on IJB-B (TAR@FAR=1e-4) over R50 TopoFR~\cite{dan2024topofr}. 
Furthermore, TransFace-L outperforms the cutting-edge method R200 AdaFace \cite{kim2022adaface} by +0.09\% on IJB-C (TAR@FAR=1e-5) and by +0.21\% on IJB-B (TAR@FAR=1e-4).
Notably, due to the limited size of the MS1MV2 training set, TransFace's representation ability is constrained, and thus it does not achieve substantial gains over SOTA competitors.
}


{On the large-scale Glint360K and WebFace42M training sets, TransFace’s powerful representation ability is fully unlocked, enabling it to clearly outperform other competitors across various backbone architectures.
Specifically, TransFace-L trained on the Glint360K dataset outperforms the SOTA competitor  LVFace~\cite{you2025lvface} by +0.14\% and +0.22\% on IJB-C (TAR@FAR=1e-4) and IJB-B (TAR@FAR=1e-4), respectively.
More importantly, when trained on the WebFace42M dataset, TransFace-L obtains the overall best results, greatly surpassing the SOTA method R200 TopoFR \cite{dan2024topofr} by +0.27\% and +0.15\% on IJB-C (TAR@FAR=1e-5) and IJB-C (TAR@FAR=1e-4), respectively.
These improved results demonstrate the superiority of our TransFace.}

\subsubsection{Results of TransFace++}

\noindent
\textbf{Results on LFW, CFP-FP, and AgeDB-30.}
We adopt MS1MV2 as the training set, and evaluate the recognition performance of our TransFace++ and compared methods on three easy benchmarks, as reported in Table~\ref{table-TransFace++}.
{Notably, in the image bytes experiments, all baseline ViT models only employ the bytes projector $\mathcal{H}$ to process image byte inputs, mapping input byte sequences longer than 37,600 into compact embeddings of length 144.}
It can be observed that regardless of the data format used, our TransFace++ models greatly outperform their corresponding baseline models, and in some cases even surpass some FR models trained with RGB tensor mentioned in Table~\ref{table-transface}.
For example, the TransFace++-S trained with TIFF encoding exhibits superior performance compared to R50 MagFace.
Moreover, when employing TIFF encoding, TransFace++-L achieves the best verification accuracy, surpassing ViT-L by 0.09\%, 1.62\%, and 1.08\% on three benchmarks, respectively.
{Although these ViT models do not surpass some cutting-edge FR methods trained with RGB tensor, these competitive results confirm that the Conv1D-based bytes mapping mechanism (from 37,600+ to 144) is capable of preserving key facial features (\emph{e.g.}, fine-grained cues essential for distinguishing between similar faces).
}

\noindent
\textbf{Results on IJB-C and IJB-B.}
We train our TransFace++ on MS1MV2 and compare it with the ViT baseline and other methods on the IJB-C benchmark, as presented in Table~\ref{table-TransFace++}.
All of our TransFace++ models achieve significant performance gains compared to their baselines.
For instance, compared to ViT-L trained with PNG encoding, its corresponding TransFace++-L achieves +2.62\% and +3.11\% improvement on IJB-C (TAR@FAR=1e-4) and IJB-B (TAR@FAR=1e-4), respectively. 
We can observe that among all data formats, training models using PNG is the most challenging because the data in PNG file contains offsets and is occasionally interrupted by bytes used for file encoding settings.

Additionally, we find that some TransFace++ models significantly outperform ViT models trained with the RGB tensor, particularly those trained with TIFF and fHWC encodings.
For example, our TransFace++-L trained with TIFF encoding surpasses the ViT-L trained with RGB tensor by +0.61\% and +0.5\% on IJB-C (TAR@FAR=1e-4) and IJB-B (TAR@FAR=1e-4), respectively.
It is worth mentioning that our TransFace++ models even work better than some ResNet-based competitors trained with RGB tensor mentioned in Table~\ref{table-transface}. For instance, our TransFace++-B trained with PNG encoding outperforms R100 DAM-CurricularFace~\cite{liu2021dam} and R100 MagFace+~\cite{meng2021magface} on two benchmarks.
Specially, when utilizing the TIFF encoding, TransFace++-L achieves the highest overall accuracy on both benchmarks. 

In addition, to further demonstrate the effectiveness of our framework in handling image bytes, we also compare the recognition performance of our TransFace++ with the classical ResNet-based FR model ArcFace~\cite{deng2019arcface} trained with different image bytes. Note that to ensure smooth input of bytes into the ArcFace model, we introduce a bytes projector $\mathcal{H}$ at the front end of the ArcFace model.
This projector $\mathcal{H}$ converts bytes into embedding and concatenates them into a feature map before feeding them into the model.
As can be seen, the compared ArcFace model exhibits poor recognition performance because they fail to effectively capture the significant correlations between image bytes. 
Furthermore, we observe that these ResNet-based models struggle to converge during training, because the input image bytes are inherently sequential data and lack the geometric structure information presented in RGB images, which severely hinders the models' understanding of image bytes. In contrast, the proposed TransFace++ effectively addresses this issue by employing the SICA module.

\subsubsection{Application of TransFace++ in Privacy-Preserving FR}
It is noteworthy that our TransFace++ framework has the potential to be applied in privacy-preserving FR tasks.
Consider a privacy-preserving camera that can mask out a portion of its pixel channels. 
The camera stores the remaining unmasked pixel channels in an array without retaining the coordinates of the pixel channels on the image sensor.
Specially, to simulate the effect of our hypothetical privacy-preserving camera, we use fHWC and fCHW image encodings as examples, predefine a fixed byte mask, and apply the mask to the fHWC and fCHW bytes, respectively.
Then, unmasked bytes are subsequently re-packed into a shorter sequence of data, which will serve as the input to the TransFace++ model.
With the help of this method, it is difficult for hackers to reconstruct users' face images at the input of the model.

\begin{table}[!t]
  \caption{Verification accuracy (\%) of different privacy-preserving FR models on LFW, CFP-FP, AgeDB-30, IJB-C and IJB-B benchmarks. 
  }
  \label{table-ppfr}
  \centering
  \resizebox{\linewidth}{!}{
  \begin{tabular}{c|c|ccc|c|c}
    \hline
     
    \multirow{2}{*}{Method} &
    \multicolumn{1}{c|}{Masking} &
    \multirow{2}{*}{LFW} &
    \multirow{2}{*}{CFP-FP} &
    \multirow{2}{*}{AgeDB-30} &
    \multicolumn{1}{c}{IJB-C} &
    \multicolumn{1}{|c}{IJB-B} 
    \\\cline{6-7}
    & Ratio &  & & & 1e-4 & 1e-4  \\
    \hline \hline

    InstaHide (RGB) \cite{huang2020instahide} & - & 96.53 & 83.20 & 79.58 & 61.88 & 69.02  \\
    
    PPFR-FD (RGB)
    \cite{wang2022privacy} & - & 99.69 & 94.85 & 97.23 & 92.93 & 94.07  \\
    
    DCTDP (RGB) \cite{ji2022privacy} & - & 99.77 & 96.97 & 97.72 & 93.29 & 94.43  \\
    AdvFace (RGB) \cite{wang2023privacy} & - & 98.45 & 92.21 & 92.57 & 70.21 & 74.39  \\
    
    MinusFace (RGB) \cite{mi2024privacy} & - & 99.78 & 96.92 & 97.57 & 93.37 & 94.70  \\
    
    \cline{1-7}
    \textbf{TransFace++-S} (fHWC) & 10\% & \textbf{99.83} & \textbf{97.96} & \textbf{97.92} & \textbf{96.18} & \textbf{95.11} \\
     \textbf{TransFace++-S} (fHWC) & 15\% & 99.83 & 97.93 & 97.88 & 96.12 & 95.07 \\

     \textbf{TransFace++-S} (fHWC) & 20\% & 99.81 & 97.85 & 97.73 & 96.04 & 94.93 \\

     \textbf{TransFace++-S} (fHWC) & 30\% & 99.80 & 97.70 & 97.64 & 95.83 & 94.81 \\
    
     \hline
     \textbf{TransFace++-S} (fCHW) & 10\% & \textbf{99.83} & 97.62 & 97.77 & 96.08 & 95.02 \\
     \textbf{TransFace++-S} (fCHW) & 15\% & 99.82 & 97.58 & 97.73 & 96.04 & 94.95 \\
     \textbf{TransFace++-S} (fCHW) & 20\% & 99.80 & 97.46 & 97.65 & 95.95 &  94.87 \\
     \textbf{TransFace++-S} (fCHW) & 30\% & 99.78 & 97.38 & 97.56 & 95.87 & 94.79 \\
    
    \hline
  \end{tabular} }
\end{table}

To showcase the potential of our TransFace++ framework in privacy-preserving FR tasks, we use MS1MV2 as the training set and evaluate the performance of our TransFace++ model across various byte mask ratios, including 10\%, 15\%, 20\% and 30\%.
Table~\ref{table-ppfr} summarizes the experimental results of our privacy-preserving camera. 
It can be observed that even at a 30\% byte mask rate, our TransFace++-S model trained with fHWC encoding still achieves high recognition performance on various benchmarks and significantly surpasses existing cutting-edge privacy-preserving FR models (e.g., AdvFace \cite{wang2023privacy}, MinusFace~\cite{mi2024privacy} and DCTDP~\cite{ji2022privacy}).

Note that in this work, we do not delve into the design of a privacy-preserving camera, as it should be studied by security experts. Instead, we simply indicate that our proposed TransFace++ framework opens up a viable path for future research on FR privacy-preserving systems.

\subsection{Analysis and Ablation Study on TransFace}

\noindent
\textbf{1) Contribution of Each Component:}
To investigate the contribution of each component in our TransFace, we employ MS1MV2 as the training set and compare TransFace-S, ViT-S (baseline), and two variants of TransFace-S on the IJB-C benchmark. The TransFace-S variants are as follows: (1) \textbf{ViT-S + SE}, the variant only introduces the SE module in the ViT-S model. (2) \textbf{ViT-S + DPAP}, based on ViT-S, the variant adds the DPAP strategy. 
{(3) \textbf{ViT-S + EHSM}, based on ViT-S, the variant adds the EHSM strategy.
}

The results in Table~\ref{tab-transface-ablation} reflect the following observations: (1) Compared to ViT-S, the accuracy of ViT-S + SE is slightly improved due to the addition of the SE module. (2) ViT-S + DPAP outperforms ViT-S + SE, indicating that
perturbing the amplitude spectrum of dominant patches can effectively alleviate the overfitting problem in ViTs.
(3) TransFace-S works better than ViT-S + DPAP, which implies the effectiveness of our EHSM strategy.

\begin{table}[t] 
\caption{Ablation study of TransFace.}
\centering
\resizebox{1.0\linewidth}{!}{
\begin{tabular}{c|c|c c c}
\hline
Training Data & Method & IJB-C(1e-6) & IJB-C(1e-5) & IJB-C(1e-4)  \\
\hline \hline 
& ViT-S &86.14 &93.40 &95.89   \\
MS1MV2 & ViT-S + SE &86.26 &93.76 &96.12   \\
& ViT-S + DPAP &86.60 &93.82 &96.30  \\
& {ViT-S + EHSM} &{86.46} & {93.85} & {96.22}  \\
& \textbf{TransFace-S} &\textbf{86.75} &\textbf{93.87} & \textbf{96.45}  \\
\hline \hline
& {ViT-S} & {88.52} & {95.24} & {96.70}   \\
& {ViT-S + SE} & {88.76} & {95.39} & {96.86}   \\
{Glint360K} & {ViT-S + DPAP} & {89.86} & {95.98} & {97.21}  \\
& {ViT-S + EHSM} & {89.92} & {95.93} & {97.15}  \\
& {\textbf{TransFace-S}} & {\textbf{90.11}} & {\textbf{96.10}} & {\textbf{97.45}}  \\
\hline
\end{tabular}
}
\label{tab-transface-ablation}
\end{table}

\noindent
\textbf{2) Comparison with Previous Data Augmentation Strategies:}
To further demonstrate the superiority of our DPAP strategy, we compare it with existing data augmentation strategies, including Random Erasing~\cite{zhong2020random}, Mixup~\cite{zhang2017mixup}, CutMix~\cite{yun2019cutmix}, RandAugment~\cite{cubuk2020randaugment}, PatchErasing~\cite{zhou2022effective}, {Fourier Domain Adaptation (\textbf{FDA})~\cite{yang2020fda}, TokenMiX~\cite{liu2022tokenmix}, and~TransMix\cite{chen2022transmix}.} 
{We train the models using MS1MV2 and Glint360K respectively, and evaluate their performance on the IJB-C benchmark.
}
As reported in Table~\ref{tab4},
compared to other strategies, our DPAP strategy can bring greater performance gain to the ViT, which benefits from the utilization of prior knowledge (\emph{i.e.}, the position of dominant patches) and the preservation of structural information of face identity.

\begin{table}[t] 
\caption{Comparison with previous data augmentation strategies.}
\centering
\resizebox{1\linewidth}{!}{
\begin{tabular}{c|c|c c c}
\hline
Training Data & Method & IJB-C(1e-6) & IJB-C(1e-5) & IJB-C(1e-4)  \\
\hline \hline 
& ViT-S &86.14 &93.40 &95.89   \\
& ViT-S + Random Erasing &83.68 &93.27 &96.06 \\
& ViT-S + RandAugment &86.24 &93.74 &96.14   \\
& ViT-S + PatchErasing & 86.26 & 93.65 & 96.11 \\
MS1MV2 & ViT-S + Mixup &85.51 &93.41 &96.08  \\
& ViT-S + CutMix &85.30 &93.53 &96.12  \\
& {ViT-S + FDA} & {86.28} & {93.38} & {96.10}  \\
& {ViT-S + TokenMix} & {82.17} & {92.87} & {95.52}  \\
& {ViT-S + TransMix} & {85.24} & {93.45} & {96.13}  \\
& \textbf{ViT-S + DPAP} &\textbf{86.60} &\textbf{93.82} & \textbf{96.30}  \\
\hline \hline
& {ViT-S} &{88.52} &{95.24} &{96.70}   \\
& {ViT-S + Random Erasing} & {86.85} & {95.16} & {96.73} \\
& {ViT-S + RandAugment} & {88.59} & {95.32} & {96.81}   \\
& {ViT-S + PatchErasing} & {88.62} & {95.35} & {96.76} \\
{Glint360K} & {ViT-S + Mixup} & {87.16} & {95.17} & {96.58}  \\
& {ViT-S + CutMix} & {86.33} & {94.82} & {96.41}   \\
& {ViT-S + FDA} & {88.73} & {95.38} & {96.75}  \\
& {ViT-S + TokenMix} & {85.64} & {94.71} & {96.47}  \\
& {ViT-S + TransMix} & {86.49} & {95.03} & {96.56}  \\
& {\textbf{ViT-S + DPAP}} & \textbf{{89.86}} & \textbf{{95.98}} & \textbf{{97.21}}  \\ 
\hline
\end{tabular}
}
\label{tab4}
\end{table}

\begin{figure}[!t]
\begin{center}   \includegraphics[width=0.9\linewidth]{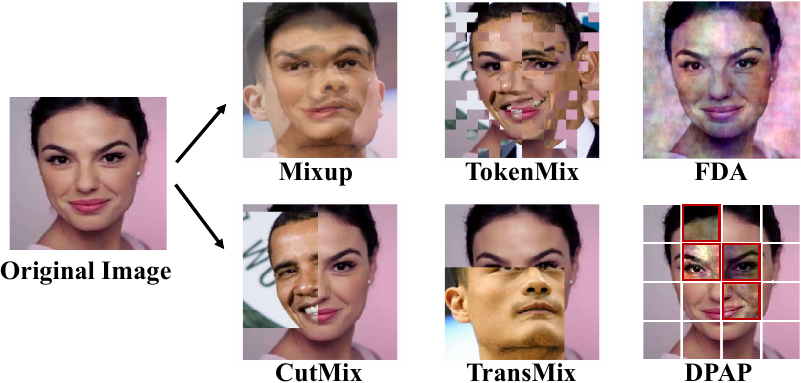}
\end{center}
   \caption{{Comparison with different data augmentation strategies.}}
\label{fig-different-aug}
\end{figure}

{To clearly demonstrate the advantages of our DPAP strategy, we also provide visual comparisons of various data augmentation strategies, as illustrated in Fig.~\ref{fig-different-aug}.
}
{Although FDA can increase sample diversity by randomly perturbing the Fourier amplitude spectrum of images, it is not well-suited for ViT-based FR frameworks. This is because, in the FR task, ViTs are highly prone to overfitting certain dominant patches (\emph{e.g.}, hair, forehead, and eyes) during training. Although FDA operates at the instance level to augment samples, it struggles to effectively address patch-level overfitting problems in ViTs.
Unlike FDA, our DPAP strategy does not simply apply random perturbations to the entire image amplitude spectrum.
Instead, DPAP employs the SE module to precisely identify the top-$K_{0}$ dominant patches that are prone to overfitting in ViTs, and then applies random perturbations specifically to the amplitude spectrum of these dominant patches, effectively augmenting the face samples at the patch level and mitigating the overfitting issue in ViTs.
}

{
As shown in Fig.~\ref{fig-different-aug}, both instance-level data augmentation methods (\emph{e.g.}, Mixup, CutMix, and TransMix) and the token-based method TokenMix are not suitable for the FR task, as they inevitably destroy key facial features and structural information of the face identity, resulting in low-quality samples that may misguide the optimization of ViTs.
In contrast, DPAP effectively preserves the local semantic coherence and identity-discriminative cues of face by perturbing the amplitude spectrum of dominant patches in the frequency domain. These subtle local features are highly valuable for the FR task.
}

{
Notably, when leveraging a larger training dataset like Glint360K, our DPAP strategy results in a more significant performance improvement for the ViT-based FR model, as it promotes the learning of more robust and generalized facial features.
However, some mixing methods, such as Mixup, CutMix, TransMix and TokenMix, experience performance degradation, as they inevitably introduce more unidentifiable face images, which generally hurts the generalization ability of the FR model.
}

\begin{figure}[!t]
\centering
\subfloat[]
{\includegraphics[width=0.5\linewidth]{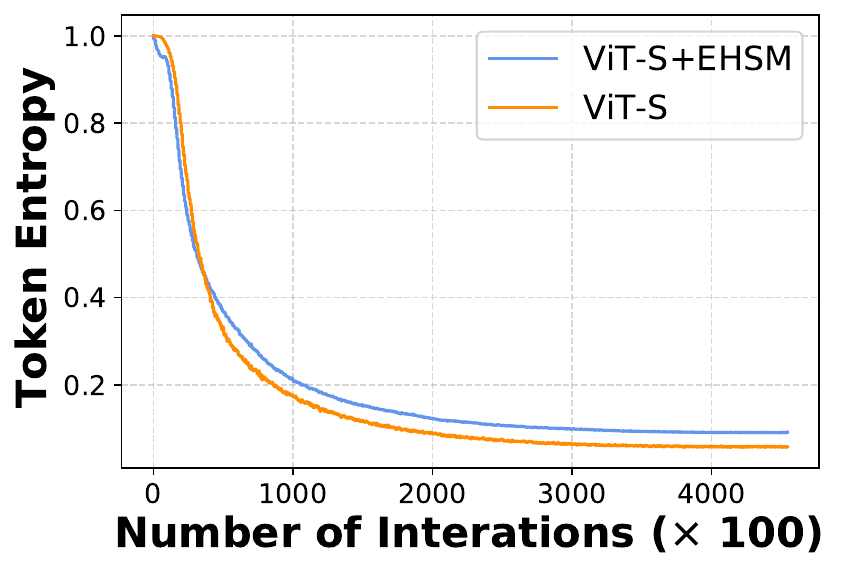}%
\label{fig-5a}}
\hfil
\subfloat[]
{\includegraphics[width=0.5\linewidth]{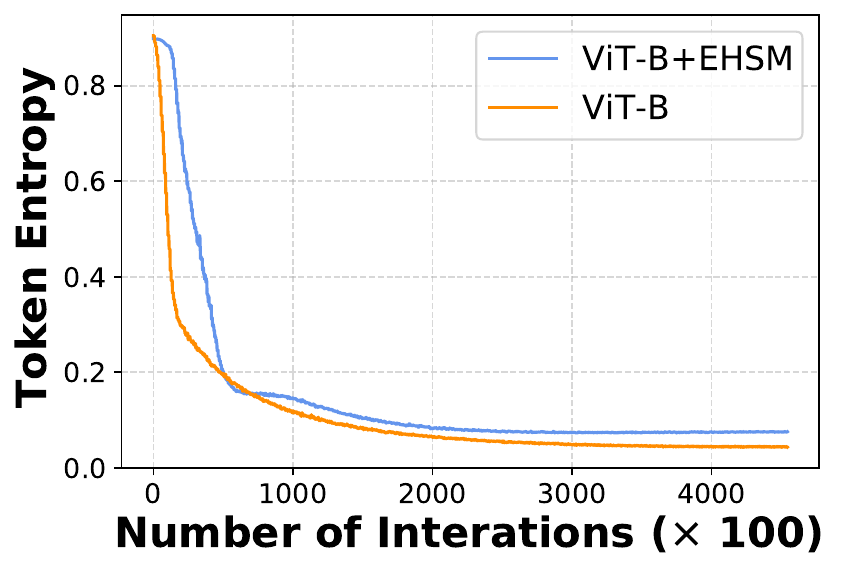}%
\label{fig-5b}}
\hfil
\subfloat[]
{\includegraphics[width=0.5\linewidth]{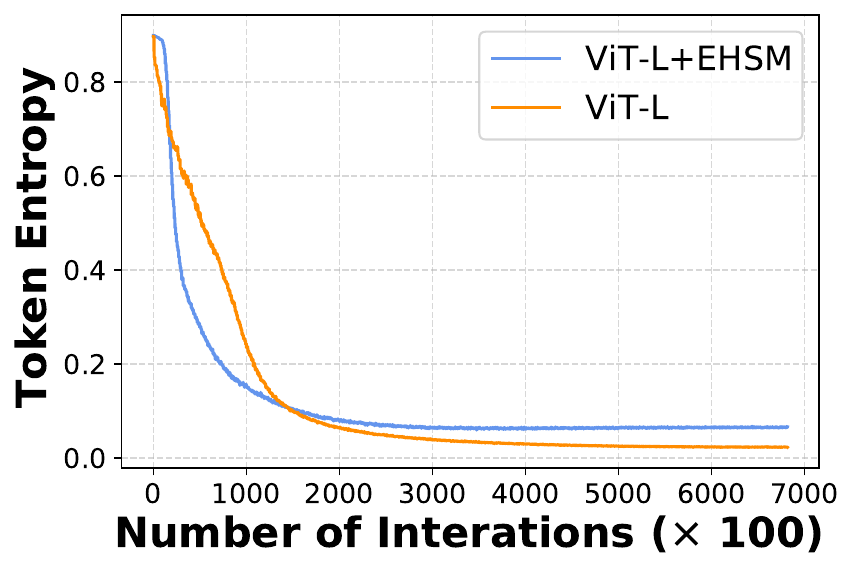}%
\label{fig-5c}}
\hfil
\subfloat[]
{\includegraphics[width=0.5\linewidth]{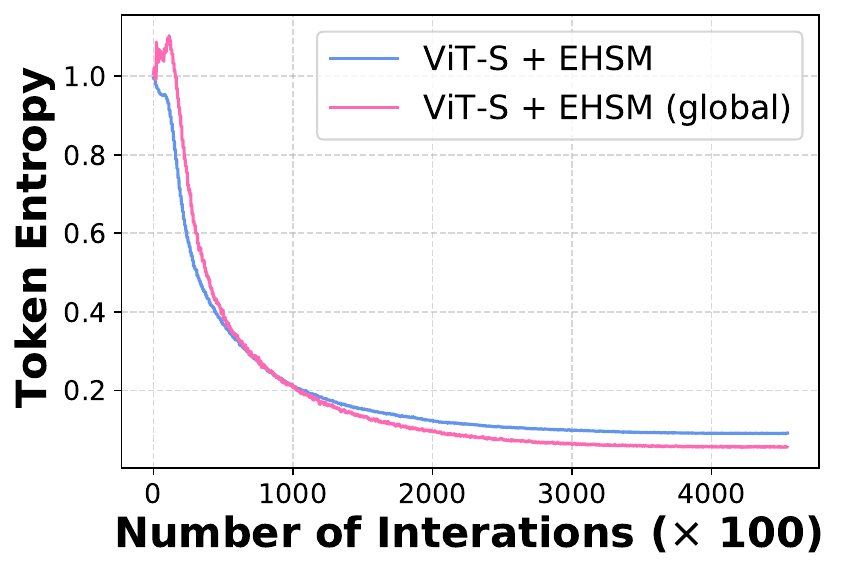}%
\label{fig-5d}}
\caption{
The trend of the average information entropy contained in the local token during training. With the help of EHSM, the face information contained in each patch is fully mined and utilized.
}
\label{fig-ehsm}
\end{figure}

\noindent
\textbf{3) Does EHSM Enhance the Feature Representation Power of Each Token?}
We investigate the trend of the average information entropy contained in the local token of ViT (baseline) and variant ViT + EHSM during training on the MS1MV2 dataset, as shown in Figs.~\ref{fig-5a},~\ref{fig-5b} and~\ref{fig-5c}.
We find that with the addition of our EHSM strategy, the token-level information becomes richer, demonstrating the superiority of the EHSM strategy in improving the feature representation power of each local token.

\noindent
\textbf{4) Effectiveness of EHSM:}
To show the superiority of EHSM in mining hard samples, we compare it with previous strategies, including AT$_{k}$ loss~\cite{fan2017learning}, MV-Softmax~\cite{wang2020mis}, Focal loss~\cite{lin2017focal}, {OHEM~\cite{shrivastava2016training}, and Token Labeling~\cite{jiang2021all}}, as reported in Table~\ref{tab-hardsample}. We can observe that the proposed EHSM strategy significantly outperforms previous hard-sample mining strategies, which indicates that our EHSM strategy can better measure sample difficulty and boost the model's performance.

{
Notably, when using MS1MV2 as the training set and ViT-S as the backbone, the model's representation power is constrained by the small size of the dataset and the simplicity of the network architecture. As a result, the performance gain from the EHSM strategy is not significant.
When employing a more complex network architecture like ViT-B and a larger-scale training dataset such as Glint360K, the performance gain achieved by the EHSM strategy becomes even more significant.
}

{
Both OHEM and Focal Loss select misclassified or hard-to-classify samples (\emph{i.e.}, hard samples) based on their loss values, allowing the model to focus more on these hard samples during training.
However, these two methods are incompatible with the ViT backbone, as they struggle to effectively leverage token-level information. As a result, applying them to mine hard samples for ViTs is a suboptimal choice.
In contrast, our EHSM strategy fully leverages the ViT architecture's strengths by comprehensively utilizing both local and global token information to precisely mine hard samples, which not only enhances the model's generalization performance, but also boosts the representation power of each local token.}
{Token Labeling aims to encourage ViT to fully exploit the rich information encoded in local image patches by introducing a machine annotator and an auxiliary loss function during training. 
However, the addition of the machine annotator and the auxiliary loss function significantly increases the computational burden. 
In contrast, our EHSM strategy encourages ViT to fully mine the fine-grained information contained in each face patch, particularly less-attended facial cues (\emph{e.g.}, lips and jaw), by maximizing the information entropy across all tokens, significantly enhancing the feature representation power of each local token.
Notably, the information entropy calculation in the EHSM strategy is straightforward and does not require additional machine annotators or loss functions, making it a lightweight solution.
}

Moreover, in order to validate the advantages of the EHSM strategy in mining hard samples by comprehensively leveraging both local and global information,
we conduct a further comparison between ViT-S{/B} + EHSM and its variant \textbf{ViT-S{/B} + EHSM (global)} that directly utilizes the entropy of the global token to measure sample difficulty. 
The results in Table~\ref{tab-hardsample} demonstrate that ViT-S{/B} + EHSM greatly outperform the variant ViT-S{/B} + EHSM (global), which indicates that fully combining the information entropy of all local tokens can more comprehensively measure the sample difficulty than only using the entropy of the global token.
Furthermore, as illustrated in Fig.~\ref{fig-5d}, compared to variant ViT-S + EHSM (global), ViT-S + EHSM can more effectively enhance the feature representation power of each local token.

\begin{table}[t]
\caption{{Comparison with previous hard sample mining strategies.}}
\centering
\resizebox{0.9\linewidth}{!}{
\begin{tabular}{c|c|c c}
\hline
Training Data & Method  & IJB-C(1e-5) & IJB-C(1e-4)  \\
\hline \hline 
& ViT-S + AT$_{k}$   &93.74  &96.03   \\
& ViT-S + MV-Softmax   &93.73  &96.08 \\
 & ViT-S + Focal loss   &93.71  &96.11 \\
MS1MV2 & {ViT-S + OHEM}   & {93.64}  & {95.97} \\
& {ViT-S + Token Labeling}   & {93.68}  & {96.05} \\
& ViT-S + EHSM (global)  &93.76 &96.13   \\
& \textbf{ViT-S + EHSM}   &\textbf{93.85}  &\textbf{96.22}   \\
\hline \hline
& {ViT-B + AT$_{k}$}   & {95.46}  & {96.89}   \\
& {ViT-B + MV-Softmax}    & {95.49}  & {96.92} \\
& {ViT-B + Focal loss}  & {95.53}  & {96.90} \\

{Glint360K} & {ViT-B + OHEM}   & {95.37}  & {96.81} \\
& {ViT-B + Token Labeling}   & {95.50}  & {96.94} \\

& {ViT-B + EHSM (global)}  & {95.88} & {97.21}   \\
& {\textbf{ViT-B + EHSM}}    &{\textbf{96.02}}  &{\textbf{97.33}}   \\
\hline
\end{tabular}
}
\label{tab-hardsample}
\end{table}

\begin{figure}[t]
\begin{center}   \includegraphics[width=1.0\linewidth]{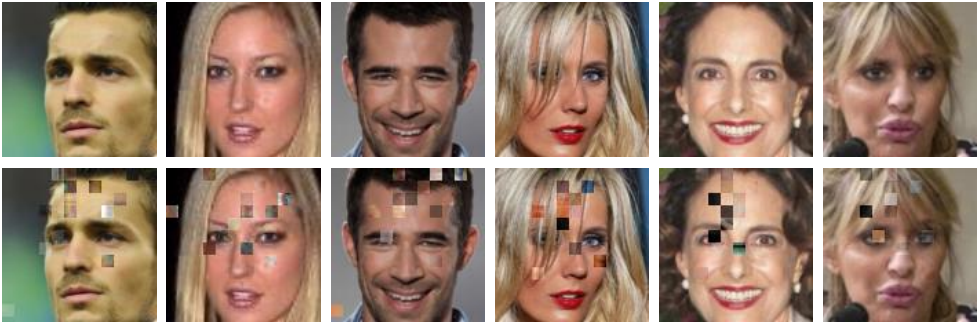}
\end{center}
   \caption{\textbf{First row:} Original training samples. \textbf{Second row:} Training samples augmented by the proposed DPAP strategy.}
\label{fig6}
\end{figure}

\noindent
\textbf{5) Visualization of DPAP:}
As shown in Fig.~\ref{fig6}, we visualize the original training samples and the samples augmented by the DPAP strategy ($K_{0}=15$) during training on MS1MV2. We can see that the dominant patches are distributed mainly near the hair, forehead, and eyes, which is consistent with our intuition. The proposed DAPA strategy effectively relieves the model from overfitting to these dominant patches by perturbing their Fourier amplitude information, which indirectly encourages ViTs to utilize the remaining facial cues (\emph{e.g.}, nose, mouth, ears, and jaw) to assist the final prediction,
significantly enhancing the model's generalization ability.

\begin{table}[t]
\caption{Ablation study of TransFace++. Training Data: MS1MV2.}
\resizebox{\linewidth}{!}{
\begin{tabular}{l|c| c| c| c}
\hline
Method & Data Format & IJB-C (1e-4) & Data Format & IJB-C (1e-4) \\
\hline \hline 
ViT-S &  & 94.74 &  & 92.88  \\
ViT-S + SE &  & 94.78 &  & 92.91  \\
ViT-S + SE + SICA & TIFF & 96.02 & PNG & 95.13  \\
ViT-S + SE + TIBC &  & 95.84 &  & 94.87  \\
\textbf{TransFace++-S} &  & \textbf{96.33} & & \textbf{95.45} \\
\hline 
\end{tabular}
} 
\label{tab5}
\end{table}

\begin{figure}[t]
\begin{center}   \includegraphics[width=1.0\linewidth]{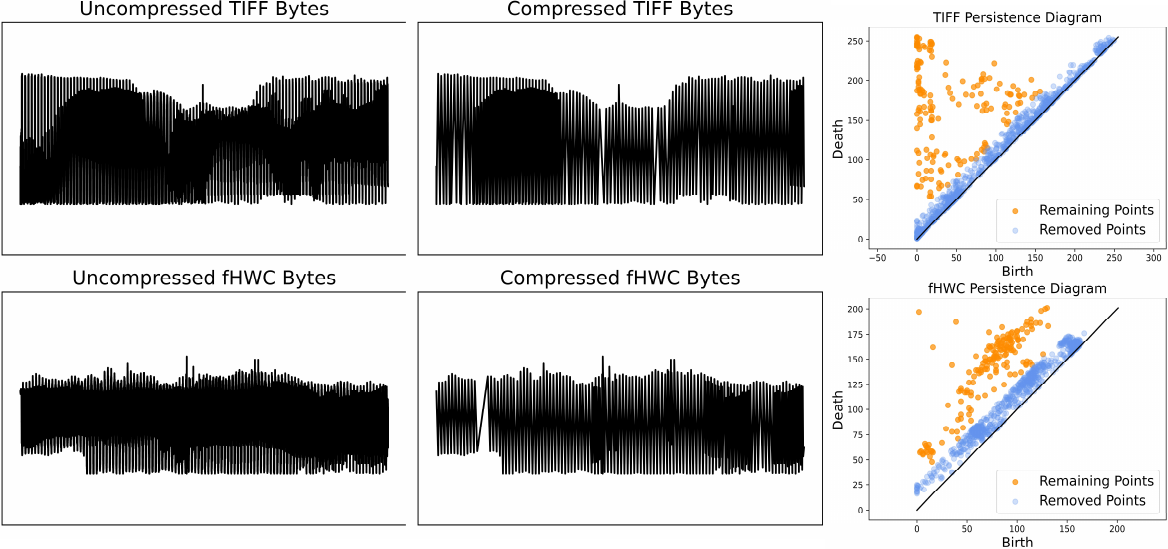}
\end{center}
   \caption{Visualization of compressed bytes features and PDs on TIFF and fHWC encodings.
   }
\label{fig3}
\end{figure}

\subsection{Analysis and Ablation Study on TransFace++}
\label{analysis-bytes}

\noindent
\textbf{1) Contribution of Each Component:}
To investigate the contribution of each component in our TransFace++, we utilize MS1MV2 with two different data formats (TIFF and PNG) as training set and compare ViT and three variants of TransFace++-S on the IJB-C benchmark. The variants of TransFace++-S are follows: (1) \textbf{ViT-S + SE}, based on ViT-S, the variant only introduces the SE module.
(2) \textbf{ViT-S + SE + SICA}, based on ViT-S + SE, the variant adds the SICA module.
(3) \textbf{ViT-S + SE + TIBC}, the variant simply introduces the TIBC strategy in the ViT-S + SE model.

The results in Table~\ref{tab5} reflect the following observations:
(1) Compared to ViT-S, the performance of ViT-S + SE is slightly improved due to the introduction of the SE module.
(2) ViT-S + SE + SICA significantly outperforms ViT-S + SE, indicating the effectiveness of our SICA module in enhancing the model's understanding of image bytes.
(3) Compared to ViT-S + SE, ViT-S + SE + TIBC shows a clear improvement in accuracy, which can be attributed to the TIBC strategy's ability to reduce information loss and preserve key facial cues. 
(4) TransFace++-S works better than all variants, implying that the SICA module is complementary to the TIBC strategy in promoting the model's understanding of image bytes.



\noindent
\textbf{{2}) Visualizaition of TIBC strategy:}
We visualize the raw image bytes (\emph{i.e.}, the original image bytes) and the compressed byte features, as depicted in Fig.~\ref{fig3}. It can be seen that our TIBC strategy excels in capturing the crucial points (\emph{i.e.}, prominent features) from the raw bytes and preserving its shape, thereby reducing information loss effectively.
Additionally, we also provide PDs for the TIFF and fHWC encodings. 
During the bytes compression process, we can observe that some low-persistence points (\emph{i.e.}, noise or redundant information) near the diagonal are discarded, while high-persistence critical points are retained. Thus, the prominent features of the bytes are effectively extracted using the TIBC strategy.

\begin{table}[!t]
\caption{Effectiveness of Probabilistic Mask Strategy.}
\centering
\resizebox{\linewidth}{!}{
\begin{tabular}{c|c| c c c }
\hline
Method & Data Format & Probability $\mu_{p}$ & IJB-C (1e-5) & IJB-C (1e-4) \\
\hline \hline 
 &  & 0 & 89.72 & 93.96 \\
 &  & 0.3 & \textbf{92.61} & \textbf{95.84}  \\
TransFace++-B & PNG & 0.5 & 91.46 & 95.33 \\
 &  & 0.7 & 90.45 & 94.57 \\
 &  & 1 & 88.68 & 93.51 \\
\hline 
\end{tabular}
}
\label{tab7}
\end{table}

\noindent
\textbf{{{3}}) Effectiveness of the Probabilistic Mask Strategy:}
To investigate the effect of the probability parameter $\mu_{p}$ on the Probabilistic Mask Strategy, we adopt the MS1MV2 dataset encoded in the PNG format to train TransFace++-B with different $\mu_{p}$ and evaluate their performance on the IJB-C benchmark, as reported in Table~\ref{tab7}. It can be observed that setting the probability $\mu_{p}$ too high will cause the model to heavily rely on the TIBC branch, thereby affecting the optimization of the bytes projector and degrading the model's performance.

\section{Conclusion}
In this paper, we rethink the limitations of existing FR paradigms in terms of efficiency, security, and precision, and propose two innovative FR frameworks to address these challenges.
To rescue the vulnerable performance of ViTs in the FR task, we develop a superior FR model called TransFace, which effectively promotes the learning of fine-grained identity features from a data-centric perspective.  
In particular, we introduce a patch-level data augmentation strategy named DPAP and a hard-sample mining strategy named EHSM. Among them, DPAP adopts a linear mix mechanism to perturb the amplitude information of dominant patches to alleviate the overfitting problem in ViTs.
EHSM fully utilizes the information entropy of multiple local tokens to measure sample difficulty, greatly enhancing the feature representation power of local tokens.
Beyond the addition of the SE module, TransFace does not introduce any significant architectural change.

Furthermore, to optimize inference efficiency and safeguard user privacy in FR systems, this paper also proposes an innovative FR framework called TransFace++, which is specifically designed to operate directly on image bytes. Concretely, TransFace++ employs an image bytes compression strategy named TIBC and a novel cross-attention module named SICA. 
TIBC leverages persistent homology to extract prominent features from the raw bytes and integrate them with byte embeddings, effectively mitigating information loss and alleviating the model's computational burden.
SICA cleverly mines the topological structure information in bytes and facilitates its interaction with byte tokens, greatly enhancing the model's understanding of image bytes.

Comprehensive experiments on various facial benchmarks verify the superiority of our TransFace and TransFace++.
We hope that our findings can shed some light on future research on ViT-based FR, as well as several relevant topics, \emph{e.g.}, FR security systems, personalized text-to-image generation model (AIGC), and 3D face reconstruction.





%





\ifCLASSOPTIONcaptionsoff

	\newpage
	\fi
	
	\bibliographystyle{IEEEtran}
	\bibliography{submission.bib}

\begin{thebibliography}{10}
\providecommand{\url}[1]{#1}
\csname url@samestyle\endcsname
\providecommand{\newblock}{\relax}
\providecommand{\bibinfo}[2]{#2}
\providecommand{\BIBentrySTDinterwordspacing}{\spaceskip=0pt\relax}
\providecommand{\BIBentryALTinterwordstretchfactor}{4}
\providecommand{\BIBentryALTinterwordspacing}{\spaceskip=\fontdimen2\font plus
\BIBentryALTinterwordstretchfactor\fontdimen3\font minus \fontdimen4\font\relax}
\providecommand{\BIBforeignlanguage}[2]{{%
\expandafter\ifx\csname l@#1\endcsname\relax
\typeout{** WARNING: IEEEtran.bst: No hyphenation pattern has been}%
\typeout{** loaded for the language `#1'. Using the pattern for}%
\typeout{** the default language instead.}%
\else
\language=\csname l@#1\endcsname
\fi
#2}}
\providecommand{\BIBdecl}{\relax}
\BIBdecl

\bibitem{cornsweet2012visual}
T.~Cornsweet, \emph{Visual perception}.\hskip 1em plus 0.5em minus 0.4em\relax Academic press, 2012.

\bibitem{dosovitskiy2020image}
A.~Dosovitskiy, L.~Beyer, A.~Kolesnikov, D.~Weissenborn, X.~Zhai, T.~Unterthiner, M.~Dehghani, M.~Minderer, G.~Heigold, S.~Gelly \emph{et~al.}, ``An image is worth 16x16 words: Transformers for image recognition at scale,'' \emph{arXiv preprint arXiv:2010.11929}, 2020.

\bibitem{touvron2021training}
H.~Touvron, M.~Cord, M.~Douze, F.~Massa, A.~Sablayrolles, and H.~J{\'e}gou, ``Training data-efficient image transformers \& distillation through attention,'' in \emph{International conference on machine learning}.\hskip 1em plus 0.5em minus 0.4em\relax PMLR, 2021, pp. 10\,347--10\,357.

\bibitem{chen2022transmix}
J.-N. Chen, S.~Sun, J.~He, P.~H. Torr, A.~Yuille, and S.~Bai, ``Transmix: Attend to mix for vision transformers,'' in \emph{Proceedings of the IEEE/CVF Conference on Computer Vision and Pattern Recognition}, 2022, pp. 12\,135--12\,144.

\bibitem{zhou2022effective}
B.~Zhou, P.~Wang, J.~Wan, Y.~Liang, and F.~Wang, ``Effective vision transformer training: A data-centric perspective,'' \emph{arXiv preprint arXiv:2209.15006}, 2022.

\bibitem{liu2021swin}
Z.~Liu, Y.~Lin, Y.~Cao, H.~Hu, Y.~Wei, Z.~Zhang, S.~Lin, and B.~Guo, ``Swin transformer: Hierarchical vision transformer using shifted windows,'' in \emph{Proceedings of the IEEE/CVF international conference on computer vision}, 2021, pp. 10\,012--10\,022.

\bibitem{zhong2021face}
Y.~Zhong and W.~Deng, ``Face transformer for recognition,'' \emph{arXiv preprint arXiv:2103.14803}, 2021.

\bibitem{xu2020learning}
K.~Xu, M.~Qin, F.~Sun, Y.~Wang, Y.-K. Chen, and F.~Ren, ``Learning in the frequency domain,'' in \emph{Proceedings of the IEEE/CVF conference on computer vision and pattern recognition}, 2020, pp. 1740--1749.

\bibitem{ehrlich2019deep}
M.~Ehrlich and L.~S. Davis, ``Deep residual learning in the jpeg transform domain,'' in \emph{Proceedings of the IEEE/CVF international conference on computer vision}, 2019, pp. 3484--3493.

\bibitem{park2023rgb}
J.~Park and J.~Johnson, ``Rgb no more: Minimally-decoded jpeg vision transformers,'' in \emph{Proceedings of the IEEE/CVF Conference on Computer Vision and Pattern Recognition}, 2023, pp. 22\,334--22\,346.

\bibitem{hortonbytes}
M.~Horton, S.~Mehta, A.~Farhadi, and M.~Rastegari, ``Bytes are all you need: Transformers operating directly on file bytes,'' \emph{Transactions on Machine Learning Research}.

\bibitem{yu2023megabyte}
L.~Yu, D.~Simig, C.~Flaherty, A.~Aghajanyan, L.~Zettlemoyer, and M.~Lewis, ``Megabyte: Predicting million-byte sequences with multiscale transformers,'' \emph{arXiv preprint arXiv:2305.07185}, 2023.

\bibitem{gersho2012vector}
A.~Gersho and R.~M. Gray, \emph{Vector quantization and signal compression}.\hskip 1em plus 0.5em minus 0.4em\relax Springer Science \& Business Media, 2012, vol. 159.

\bibitem{edelsbrunner2006persistence}
H.~Edelsbrunner, D.~Morozov, and V.~Pascucci, ``Persistence-sensitive simplification functions on 2-manifolds,'' in \emph{Proceedings of the twenty-second annual symposium on Computational geometry}, 2006, pp. 127--134.

\bibitem{zomorodian2004computing}
A.~Zomorodian and G.~Carlsson, ``Computing persistent homology,'' in \emph{Proceedings of the twentieth annual symposium on Computational geometry}, 2004, pp. 347--356.

\bibitem{wu2021rethinking}
K.~Wu, H.~Peng, M.~Chen, J.~Fu, and H.~Chao, ``Rethinking and improving relative position encoding for vision transformer,'' in \emph{Proceedings of the IEEE/CVF International Conference on Computer Vision}, 2021, pp. 10\,033--10\,041.

\bibitem{carlsson2009topology}
G.~Carlsson, ``Topology and data,'' \emph{Bulletin of the American Mathematical Society}, vol.~46, no.~2, pp. 255--308, 2009.

\bibitem{moor2020topological}
M.~Moor, M.~Horn, B.~Rieck, and K.~Borgwardt, ``Topological autoencoders,'' in \emph{International conference on machine learning}.\hskip 1em plus 0.5em minus 0.4em\relax PMLR, 2020, pp. 7045--7054.

\bibitem{dan2023transface}
J.~Dan, Y.~Liu, H.~Xie, J.~Deng, H.~Xie, X.~Xie, and B.~Sun, ``Transface: Calibrating transformer training for face recognition from a data-centric perspective,'' in \emph{Proceedings of the IEEE/CVF International Conference on Computer Vision}, 2023, pp. 20\,642--20\,653.

\bibitem{li2024transformer}
X.~Li, H.~Ding, H.~Yuan, W.~Zhang, J.~Pang, G.~Cheng, K.~Chen, Z.~Liu, and C.~C. Loy, ``Transformer-based visual segmentation: A survey,'' \emph{IEEE transactions on pattern analysis and machine intelligence}, 2024.

\bibitem{hao2025simple}
C.~Hao, Z.~Yu, X.~Liu, J.~Xu, H.~Yue, and J.~Yang, ``A simple yet effective network based on vision transformer for camouflaged object and salient object detection,'' \emph{IEEE Transactions on Image Processing}, 2025.

\bibitem{kim2024learning}
M.~Kim, P.~H. Seo, C.~Schmid, and M.~Cho, ``Learning correlation structures for vision transformers,'' in \emph{Proceedings of the IEEE/CVF conference on computer vision and pattern recognition}, 2024, pp. 18\,941--18\,951.

\bibitem{han2021transformer}
K.~Han, A.~Xiao, E.~Wu, J.~Guo, C.~Xu, and Y.~Wang, ``Transformer in transformer,'' \emph{Advances in Neural Information Processing Systems}, vol.~34, pp. 15\,908--15\,919, 2021.

\bibitem{liu2022tokenmix}
J.~Liu, B.~Liu, H.~Zhou, H.~Li, and Y.~Liu, ``Tokenmix: Rethinking image mixing for data augmentation in vision transformers,'' in \emph{European conference on computer vision}.\hskip 1em plus 0.5em minus 0.4em\relax Springer, 2022, pp. 455--471.

\bibitem{schroff2015facenet}
F.~Schroff, D.~Kalenichenko, and J.~Philbin, ``Facenet: A unified embedding for face recognition and clustering,'' in \emph{Proceedings of the IEEE conference on computer vision and pattern recognition}, 2015, pp. 815--823.

\bibitem{sohn2016improved}
K.~Sohn, ``Improved deep metric learning with multi-class n-pair loss objective,'' \emph{Advances in neural information processing systems}, vol.~29, 2016.

\bibitem{wen2016discriminative}
Y.~Wen, K.~Zhang, Z.~Li, and Y.~Qiao, ``A discriminative feature learning approach for deep face recognition,'' in \emph{Computer Vision--ECCV 2016: 14th European Conference, Amsterdam, The Netherlands, October 11--14, 2016, Proceedings, Part VII 14}.\hskip 1em plus 0.5em minus 0.4em\relax Springer, 2016, pp. 499--515.

\bibitem{deng2019arcface}
J.~Deng, J.~Guo, N.~Xue, and S.~Zafeiriou, ``Arcface: Additive angular margin loss for deep face recognition,'' in \emph{Proceedings of the IEEE/CVF conference on computer vision and pattern recognition}, 2019, pp. 4690--4699.

\bibitem{wang2018cosface}
H.~Wang, Y.~Wang, Z.~Zhou, X.~Ji, D.~Gong, J.~Zhou, Z.~Li, and W.~Liu, ``Cosface: Large margin cosine loss for deep face recognition,'' in \emph{Proceedings of the IEEE conference on computer vision and pattern recognition}, 2018, pp. 5265--5274.

\bibitem{zhou2023uniface}
J.~Zhou, X.~Jia, Q.~Li, L.~Shen, and J.~Duan, ``Uniface: Unified cross-entropy loss for deep face recognition,'' in \emph{Proceedings of the IEEE/CVF International Conference on Computer Vision}, 2023, pp. 20\,730--20\,739.

\bibitem{meng2021magface}
Q.~Meng, S.~Zhao, Z.~Huang, and F.~Zhou, ``Magface: A universal representation for face recognition and quality assessment,'' in \emph{Proceedings of the IEEE/CVF Conference on Computer Vision and Pattern Recognition}, 2021, pp. 14\,225--14\,234.

\bibitem{li2021spherical}
S.~Li, J.~Xu, X.~Xu, P.~Shen, S.~Li, and B.~Hooi, ``Spherical confidence learning for face recognition,'' in \emph{Proceedings of the IEEE/CVF Conference on Computer Vision and Pattern Recognition}, 2021, pp. 15\,629--15\,637.

\bibitem{kim2022adaface}
M.~Kim, A.~K. Jain, and X.~Liu, ``Adaface: Quality adaptive margin for face recognition,'' in \emph{Proceedings of the IEEE/CVF Conference on Computer Vision and Pattern Recognition}, 2022, pp. 18\,750--18\,759.

\bibitem{duan2019uniformface}
Y.~Duan, J.~Lu, and J.~Zhou, ``Uniformface: Learning deep equidistributed representation for face recognition,'' in \emph{Proceedings of the IEEE/CVF Conference on Computer Vision and Pattern Recognition}, 2019, pp. 3415--3424.

\bibitem{huang2020curricularface}
Y.~Huang, Y.~Wang, Y.~Tai, X.~Liu, P.~Shen, S.~Li, J.~Li, and F.~Huang, ``Curricularface: adaptive curriculum learning loss for deep face recognition,'' in \emph{proceedings of the IEEE/CVF conference on computer vision and pattern recognition}, 2020, pp. 5901--5910.

\bibitem{dan2024topofr}
J.~Dan, Y.~Liu, J.~Deng, H.~Xie, S.~Li, B.~Sun, and S.~Luo, ``Topofr: A closer look at topology alignment on face recognition,'' \emph{Advances in Neural Information Processing Systems}, vol.~37, pp. 37\,213--37\,240, 2024.

\bibitem{an2021partial}
X.~An, X.~Zhu, Y.~Gao, Y.~Xiao, Y.~Zhao, Z.~Feng, L.~Wu, B.~Qin, M.~Zhang, D.~Zhang \emph{et~al.}, ``Partial fc: Training 10 million identities on a single machine,'' in \emph{Proceedings of the IEEE/CVF International Conference on Computer Vision}, 2021, pp. 1445--1449.

\bibitem{an2022killing}
X.~An, J.~Deng, J.~Guo, Z.~Feng, X.~Zhu, J.~Yang, and T.~Liu, ``Killing two birds with one stone: Efficient and robust training of face recognition cnns by partial fc,'' in \emph{Proceedings of the IEEE/CVF Conference on Computer Vision and Pattern Recognition}, 2022, pp. 4042--4051.

\bibitem{caldeira2024mst}
E.~Caldeira, J.~S. Cardoso, A.~F. Sequeira, and P.~C. Neto, ``Mst-kd: Multiple specialized teachers knowledge distillation for fair face recognition,'' in \emph{European Conference on Computer Vision}.\hskip 1em plus 0.5em minus 0.4em\relax Springer, 2024, pp. 211--228.

\bibitem{boutros2024adadistill}
F.~Boutros, V.~{\v{S}}truc, and N.~Damer, ``Adadistill: Adaptive knowledge distillation for deep face recognition,'' in \emph{European Conference on Computer Vision}.\hskip 1em plus 0.5em minus 0.4em\relax Springer, 2024, pp. 163--182.

\bibitem{qin2023swinface}
L.~Qin, M.~Wang, C.~Deng, K.~Wang, X.~Chen, J.~Hu, and W.~Deng, ``Swinface: a multi-task transformer for face recognition, expression recognition, age estimation and attribute estimation,'' \emph{IEEE Transactions on Circuits and Systems for Video Technology}, vol.~34, no.~4, pp. 2223--2234, 2023.

\bibitem{gueguen2018faster}
L.~Gueguen, A.~Sergeev, B.~Kadlec, R.~Liu, and J.~Yosinski, ``Faster neural networks straight from jpeg,'' \emph{Advances in Neural Information Processing Systems}, vol.~31, 2018.

\bibitem{adams2017persistence}
H.~Adams, T.~Emerson, M.~Kirby, R.~Neville, C.~Peterson, P.~Shipman, S.~Chepushtanova, E.~Hanson, F.~Motta, and L.~Ziegelmeier, ``Persistence images: A stable vector representation of persistent homology,'' \emph{Journal of Machine Learning Research}, vol.~18, 2017.

\bibitem{zomorodian2010fast}
A.~Zomorodian, ``Fast construction of the vietoris-rips complex,'' \emph{Computers \& Graphics}, vol.~34, no.~3, pp. 263--271, 2010.

\bibitem{som2018perturbation}
A.~Som, K.~Thopalli, K.~N. Ramamurthy, V.~Venkataraman, A.~Shukla, and P.~Turaga, ``Perturbation robust representations of topological persistence diagrams,'' in \emph{Proceedings of the European Conference on Computer Vision (ECCV)}, 2018, pp. 617--635.

\bibitem{clark2015pillow}
A.~Clark \emph{et~al.}, ``Pillow (pil fork) documentation,'' \emph{readthedocs}, 2015.

\bibitem{boutell1997png}
T.~Boutell, ``Png (portable network graphics) specification version 1.0,'' Tech. Rep., 1997.

\bibitem{8578843}
J.~Hu, L.~Shen, and G.~Sun, ``Squeeze-and-excitation networks,'' in \emph{2018 IEEE/CVF Conference on Computer Vision and Pattern Recognition}, 2018, pp. 7132--7141.

\bibitem{zhong2020random}
Z.~Zhong, L.~Zheng, G.~Kang, S.~Li, and Y.~Yang, ``Random erasing data augmentation,'' in \emph{Proceedings of the AAAI conference on artificial intelligence}, vol.~34, no.~07, 2020, pp. 13\,001--13\,008.

\bibitem{zhang2017mixup}
H.~Zhang, M.~Cisse, Y.~N. Dauphin, and D.~Lopez-Paz, ``mixup: Beyond empirical risk minimization,'' \emph{arXiv preprint arXiv:1710.09412}, 2017.

\bibitem{yun2019cutmix}
S.~Yun, D.~Han, S.~J. Oh, S.~Chun, J.~Choe, and Y.~Yoo, ``Cutmix: Regularization strategy to train strong classifiers with localizable features,'' in \emph{Proceedings of the IEEE/CVF international conference on computer vision}, 2019, pp. 6023--6032.

\bibitem{cubuk2020randaugment}
E.~D. Cubuk, B.~Zoph, J.~Shlens, and Q.~V. Le, ``Randaugment: Practical automated data augmentation with a reduced search space,'' in \emph{Proceedings of the IEEE/CVF conference on computer vision and pattern recognition workshops}, 2020, pp. 702--703.

\bibitem{lin2017focal}
T.-Y. Lin, P.~Goyal, R.~Girshick, K.~He, and P.~Doll{\'a}r, ``Focal loss for dense object detection,'' in \emph{Proceedings of the IEEE international conference on computer vision}, 2017, pp. 2980--2988.

\bibitem{huang2020improving}
Y.~Huang, P.~Shen, Y.~Tai, S.~Li, X.~Liu, J.~Li, F.~Huang, and R.~Ji, ``Improving face recognition from hard samples via distribution distillation loss,'' in \emph{Computer Vision--ECCV 2020: 16th European Conference, Glasgow, UK, August 23--28, 2020, Proceedings, Part XXX 16}.\hskip 1em plus 0.5em minus 0.4em\relax Springer, 2020, pp. 138--154.

\bibitem{fan2017learning}
Y.~Fan, S.~Lyu, Y.~Ying, and B.~Hu, ``Learning with average top-k loss,'' \emph{Advances in neural information processing systems}, vol.~30, 2017.

\bibitem{shrivastava2016training}
A.~Shrivastava, A.~Gupta, and R.~Girshick, ``Training region-based object detectors with online hard example mining,'' in \emph{Proceedings of the IEEE conference on computer vision and pattern recognition}, 2016, pp. 761--769.

\bibitem{piotrowski1982demonstration}
L.~N. Piotrowski and F.~W. Campbell, ``A demonstration of the visual importance and flexibility of spatial-frequency amplitude and phase,'' \emph{Perception}, vol.~11, no.~3, pp. 337--346, 1982.

\bibitem{yang2020fda}
Y.~Yang and S.~Soatto, ``Fda: Fourier domain adaptation for semantic segmentation,'' in \emph{Proceedings of the IEEE/CVF Conference on Computer Vision and Pattern Recognition}, 2020, pp. 4085--4095.

\bibitem{brigham1967fast}
E.~O. Brigham and R.~Morrow, ``The fast fourier transform,'' \emph{IEEE spectrum}, vol.~4, no.~12, pp. 63--70, 1967.

\bibitem{wang2020mis}
X.~Wang, S.~Zhang, S.~Wang, T.~Fu, H.~Shi, and T.~Mei, ``Mis-classified vector guided softmax loss for face recognition,'' in \emph{Proceedings of the AAAI Conference on Artificial Intelligence}, vol.~34, no.~07, 2020, pp. 12\,241--12\,248.

\bibitem{cover1999elements}
T.~M. Cover, \emph{Elements of information theory}.\hskip 1em plus 0.5em minus 0.4em\relax John Wiley \& Sons, 1999.

\bibitem{maze2018iarpa}
B.~Maze, J.~Adams, J.~A. Duncan, N.~Kalka, T.~Miller, C.~Otto, A.~K. Jain, W.~T. Niggel, J.~Anderson, J.~Cheney \emph{et~al.}, ``Iarpa janus benchmark-c: Face dataset and protocol,'' in \emph{2018 international conference on biometrics (ICB)}.\hskip 1em plus 0.5em minus 0.4em\relax IEEE, 2018, pp. 158--165.

\bibitem{fang2022msg}
J.~Fang, L.~Xie, X.~Wang, X.~Zhang, W.~Liu, and Q.~Tian, ``Msg-transformer: Exchanging local spatial information by manipulating messenger tokens,'' in \emph{Proceedings of the IEEE/CVF Conference on Computer Vision and Pattern Recognition}, 2022, pp. 12\,063--12\,072.

\bibitem{edelsbrunner2002topological}
Edelsbrunner, Letscher, and Zomorodian, ``Topological persistence and simplification,'' \emph{Discrete \& Computational Geometry}, vol.~28, pp. 511--533, 2002.

\bibitem{chazal2009proximity}
F.~Chazal, D.~Cohen-Steiner, M.~Glisse, L.~J. Guibas, and S.~Y. Oudot, ``Proximity of persistence modules and their diagrams,'' in \emph{Proceedings of the twenty-fifth annual symposium on Computational geometry}, 2009, pp. 237--246.

\bibitem{milnor1963morse}
J.~W. Milnor, \emph{Morse theory}.\hskip 1em plus 0.5em minus 0.4em\relax Princeton university press, 1963, no.~51.

\bibitem{karim2011wavelet}
S.~A.~A. Karim, M.~H. Kamarudin, B.~A. Karim, M.~K. Hasan, and J.~Sulaiman, ``Wavelet transform and fast fourier transform for signal compression: A comparative study,'' in \emph{2011 International Conference on Electronic Devices, Systems and Applications (ICEDSA)}.\hskip 1em plus 0.5em minus 0.4em\relax IEEE, 2011, pp. 280--285.

\bibitem{gilbert2014recent}
A.~C. Gilbert, P.~Indyk, M.~Iwen, and L.~Schmidt, ``Recent developments in the sparse fourier transform: A compressed fourier transform for big data,'' \emph{IEEE Signal Processing Magazine}, vol.~31, no.~5, pp. 91--100, 2014.

\bibitem{kim2020broadface}
Y.~Kim, W.~Park, and J.~Shin, ``Broadface: Looking at tens of thousands of people at once for face recognition,'' in \emph{Computer Vision--ECCV 2020: 16th European Conference, Glasgow, UK, August 23--28, 2020, Proceedings, Part IX}.\hskip 1em plus 0.5em minus 0.4em\relax Springer, 2020, pp. 536--552.

\bibitem{liu2021dam}
J.~Liu, Y.~Wu, Y.~Wu, C.~Li, X.~Hu, D.~Liang, and M.~Wang, ``Dam: Discrepancy alignment metric for face recognition,'' in \emph{Proceedings of the IEEE/CVF International Conference on Computer Vision}, 2021, pp. 3814--3823.

\bibitem{Boutros_2022_CVPR}
F.~Boutros, N.~Damer, F.~Kirchbuchner, and A.~Kuijper, ``Elasticface: Elastic margin loss for deep face recognition,'' in \emph{Proceedings of the IEEE/CVF Conference on Computer Vision and Pattern Recognition (CVPR) Workshops}, June 2022, pp. 1578--1587.

\bibitem{you2025lvface}
J.~You, S.~Li, Y.~Sun, J.~Wei, M.~Guo, C.~Feng, and J.~Ran, ``{LVFace}: Progressive cluster optimization for large vision models in face recognition,'' in \emph{ICCV}, 2025.

\bibitem{li2023unitsface}
Q.~Li, X.~Jia, J.~Zhou, L.~Shen, and J.~Duan, ``Unitsface: Unified threshold integrated sample-to-sample loss for face recognition,'' in \emph{Thirty-seventh Conference on Neural Information Processing Systems}, 2023.

\bibitem{zhu2021webface260m}
Z.~Zhu, G.~Huang, J.~Deng, Y.~Ye, J.~Huang, X.~Chen, J.~Zhu, T.~Yang, J.~Lu, D.~Du \emph{et~al.}, ``Webface260m: A benchmark unveiling the power of million-scale deep face recognition,'' in \emph{Proceedings of the IEEE/CVF Conference on Computer Vision and Pattern Recognition}, 2021, pp. 10\,492--10\,502.

\bibitem{boutros2022sface}
F.~Boutros, M.~Huber, P.~Siebke, T.~Rieber, and N.~Damer, ``Sface: Privacy-friendly and accurate face recognition using synthetic data,'' in \emph{2022 IEEE International Joint Conference on Biometrics (IJCB)}.\hskip 1em plus 0.5em minus 0.4em\relax IEEE, 2022, pp. 1--11.

\bibitem{huang2008labeled}
G.~B. Huang, M.~Mattar, T.~Berg, and E.~Learned-Miller, ``Labeled faces in the wild: A database forstudying face recognition in unconstrained environments,'' in \emph{Workshop on faces in'Real-Life'Images: detection, alignment, and recognition}, 2008.

\bibitem{moschoglou2017agedb}
S.~Moschoglou, A.~Papaioannou, C.~Sagonas, J.~Deng, I.~Kotsia, and S.~Zafeiriou, ``Agedb: the first manually collected, in-the-wild age database,'' in \emph{proceedings of the IEEE conference on computer vision and pattern recognition workshops}, 2017, pp. 51--59.

\bibitem{sengupta2016frontal}
S.~Sengupta, J.-C. Chen, C.~Castillo, V.~M. Patel, R.~Chellappa, and D.~W. Jacobs, ``Frontal to profile face verification in the wild,'' in \emph{2016 IEEE winter conference on applications of computer vision (WACV)}.\hskip 1em plus 0.5em minus 0.4em\relax IEEE, 2016, pp. 1--9.

\bibitem{whitelam2017iarpa}
C.~Whitelam, E.~Taborsky, A.~Blanton, B.~Maze, J.~Adams, T.~Miller, N.~Kalka, A.~K. Jain, J.~A. Duncan, K.~Allen \emph{et~al.}, ``Iarpa janus benchmark-b face dataset,'' in \emph{proceedings of the IEEE conference on computer vision and pattern recognition workshops}, 2017, pp. 90--98.

\bibitem{zheng2018cross}
T.~Zheng and W.~Deng, ``Cross-pose lfw: A database for studying cross-pose face recognition in unconstrained environments,'' \emph{Beijing University of Posts and Telecommunications, Tech. Rep}, vol.~5, no.~7, p.~5, 2018.

\bibitem{zheng2017cross}
T.~Zheng, W.~Deng, and J.~Hu, ``Cross-age lfw: A database for studying cross-age face recognition in unconstrained environments,'' \emph{arXiv preprint arXiv:1708.08197}, 2017.

\bibitem{cheng2019low}
Z.~Cheng, X.~Zhu, and S.~Gong, ``Low-resolution face recognition,'' in \emph{Computer Vision--ACCV 2018: 14th Asian Conference on Computer Vision, Perth, Australia, December 2--6, 2018, Revised Selected Papers, Part III 14}.\hskip 1em plus 0.5em minus 0.4em\relax Springer, 2019, pp. 605--621.

\bibitem{kemelmacher2016megaface}
I.~Kemelmacher-Shlizerman, S.~M. Seitz, D.~Miller, and E.~Brossard, ``The megaface benchmark: 1 million faces for recognition at scale,'' in \emph{Proceedings of the IEEE conference on computer vision and pattern recognition}, 2016, pp. 4873--4882.

\bibitem{wang2019racial}
M.~Wang, W.~Deng, J.~Hu, X.~Tao, and Y.~Huang, ``Racial faces in the wild: Reducing racial bias by information maximization adaptation network,'' in \emph{Proceedings of the ieee/cvf international conference on computer vision}, 2019, pp. 692--702.

\bibitem{deng2021masked}
J.~Deng, J.~Guo, X.~An, Z.~Zhu, and S.~Zafeiriou, ``Masked face recognition challenge: The insightface track report,'' in \emph{Proceedings of the IEEE/CVF International Conference on Computer Vision}, 2021, pp. 1437--1444.

\bibitem{huang2020instahide}
Y.~Huang, Z.~Song, K.~Li, and S.~Arora, ``Instahide: Instance-hiding schemes for private distributed learning,'' in \emph{International conference on machine learning}.\hskip 1em plus 0.5em minus 0.4em\relax PMLR, 2020, pp. 4507--4518.

\bibitem{wang2022privacy}
Y.~Wang, J.~Liu, M.~Luo, L.~Yang, and L.~Wang, ``Privacy-preserving face recognition in the frequency domain,'' in \emph{Proceedings of the AAAI Conference on Artificial Intelligence}, vol.~36, no.~3, 2022, pp. 2558--2566.

\bibitem{ji2022privacy}
J.~Ji, H.~Wang, Y.~Huang, J.~Wu, X.~Xu, S.~Ding, S.~Zhang, L.~Cao, and R.~Ji, ``Privacy-preserving face recognition with learnable privacy budgets in frequency domain,'' in \emph{European Conference on Computer Vision}.\hskip 1em plus 0.5em minus 0.4em\relax Springer, 2022, pp. 475--491.

\bibitem{wang2023privacy}
Z.~Wang, H.~Wang, S.~Jin, W.~Zhang, J.~Hu, Y.~Wang, P.~Sun, W.~Yuan, K.~Liu, and K.~Ren, ``Privacy-preserving adversarial facial features,'' in \emph{Proceedings of the IEEE/CVF Conference on Computer Vision and Pattern Recognition}, 2023, pp. 8212--8221.

\bibitem{mi2024privacy}
Y.~Mi, Z.~Zhong, Y.~Huang, J.~Ji, J.~Xu, J.~Wang, S.~Wang, S.~Ding, and S.~Zhou, ``Privacy-preserving face recognition using trainable feature subtraction,'' \emph{arXiv preprint arXiv:2403.12457}, 2024.

\bibitem{jiang2021all}
Z.-H. Jiang, Q.~Hou, L.~Yuan, D.~Zhou, Y.~Shi, X.~Jin, A.~Wang, and J.~Feng, ``All tokens matter: Token labeling for training better vision transformers,'' \emph{Advances in neural information processing systems}, vol.~34, pp. 18\,590--18\,602, 2021.

\end{thebibliography}

\begin{IEEEbiography}[{\includegraphics[width=1in,height=1.25in,clip,keepaspectratio]{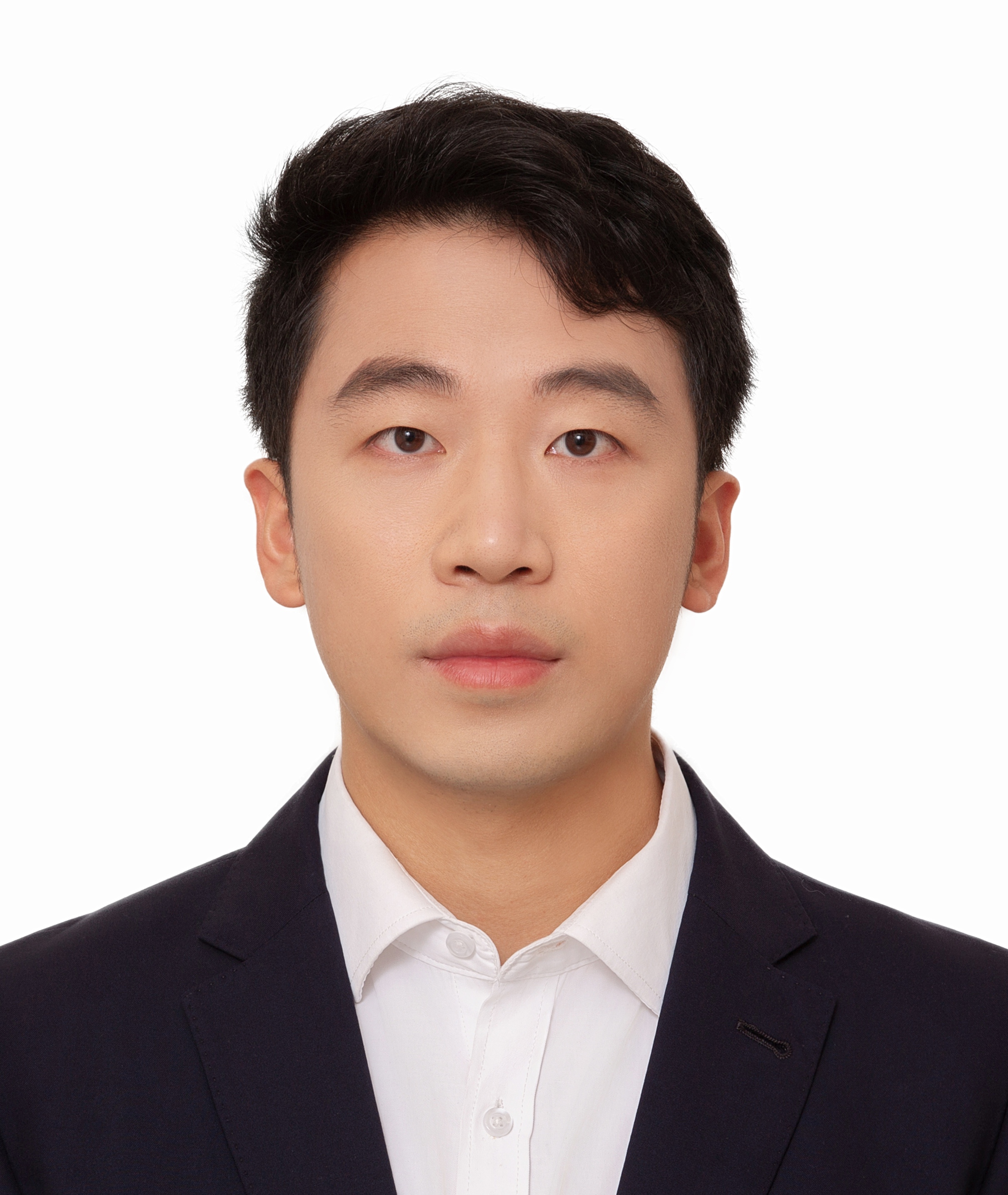}}]{Jun Dan} is a Ph.D. candidate at the College of Information Science \& Electronic Engineering, Zhejiang University, China. 
He received the B.Eng. degree in Electronic Information Engineering from Chongqing University, China, in 2020. 
He serves as a reviewer for prestigious computer vision conferences and journals, including CVPR, ECCV, ICCV, TIP, ICLR, NeurIPS, and ACM MM.
His research interests include transfer learning, face recognition, and LMMs.
\end{IEEEbiography}

\begin{IEEEbiography}[{\includegraphics[width=1in,height=1.25in,clip,keepaspectratio]{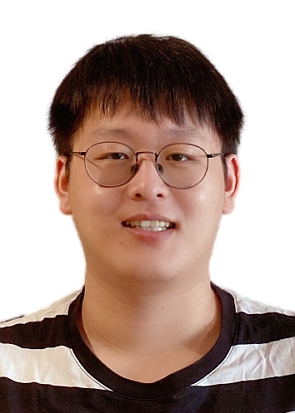}}]{Yang Liu} is a Ph.D. candidate at the Department of Engineering, King’s College London.
He received his M.Sc. degree in automation from North China Electric Power University, China in 2020. 
He is a reviewer in prestigious computer vision conferences including CVPR, ICCV, ICML, ICLR, and NeurIPS.
His main research interest is deep face perception and understanding and has published over 4 first-author face-related  papers in CVPR and ICLR.
\end{IEEEbiography}



\begin{IEEEbiography}[{\includegraphics[width=1in,height=1.25in,clip, keepaspectratio]{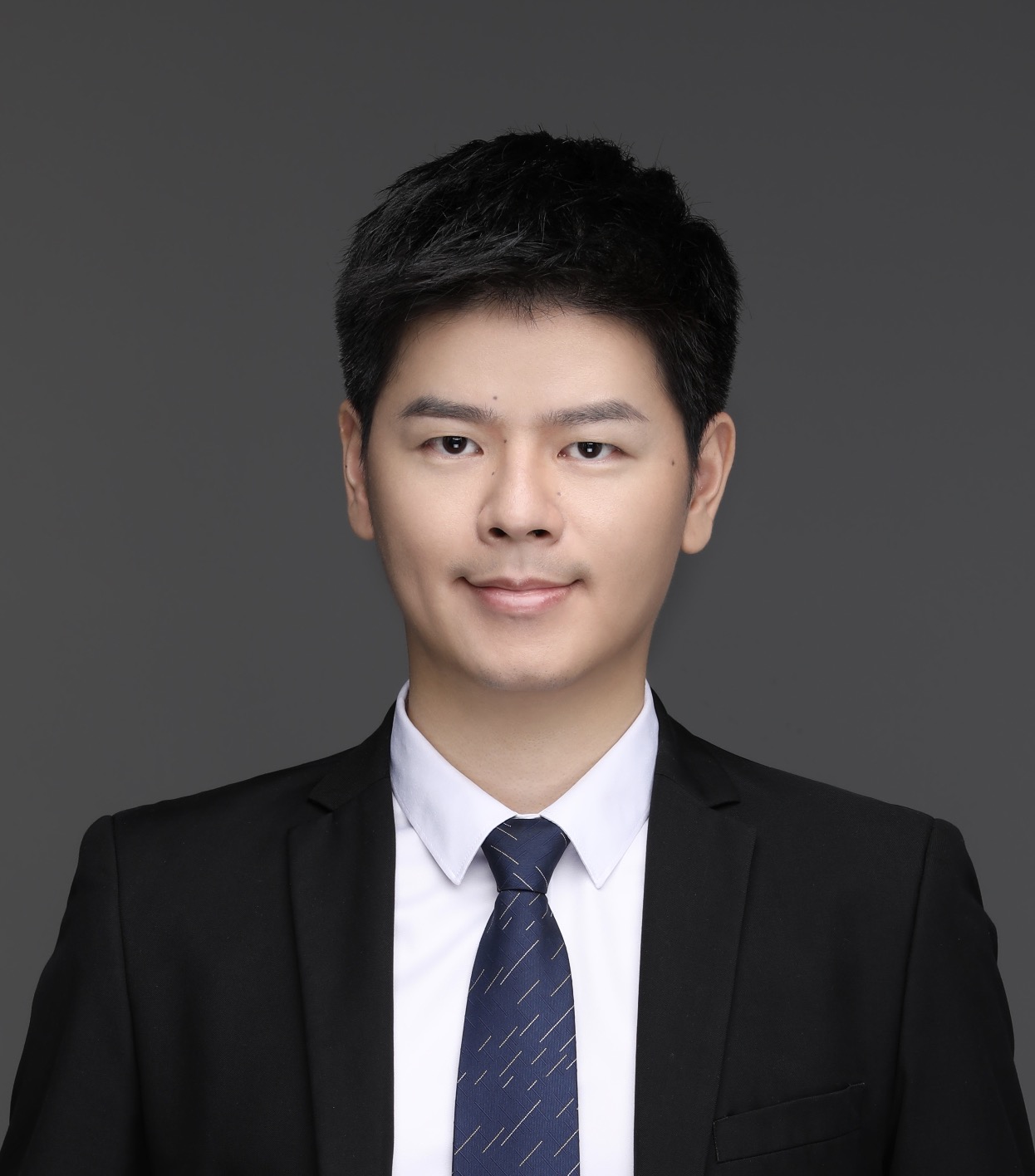}}]{Baigui Sun}
is the AI Lead at Wolf 1069 b Lab, Sany Group, where he drives research in embodied intelligence, AIGC, and the applications of MLLMs/LLMs.
Previously, he was an early team member at Alibaba DAMO Academy and Tongyi Lab, contributing to the development of many widely-used AI applications (e.g., Pailitao Image Search, Taobao DeepCTR Search, Content Moderation, Face Recognition, FaceChain, and ModelScope). 
Baigui holds an M.S. from Zhejiang University and has authored research cited over 2,000 times.
He is the founding member of the popular open-source project FaceChain and serves as a reviewer for prestigious computer vision conferences and journals, including CVPR, ICCV, ICML, NeurIPS, ICLR, and TPAMI.
\end{IEEEbiography}

\begin{IEEEbiography}[{\includegraphics[width=1in,height=1.25in,clip,keepaspectratio]{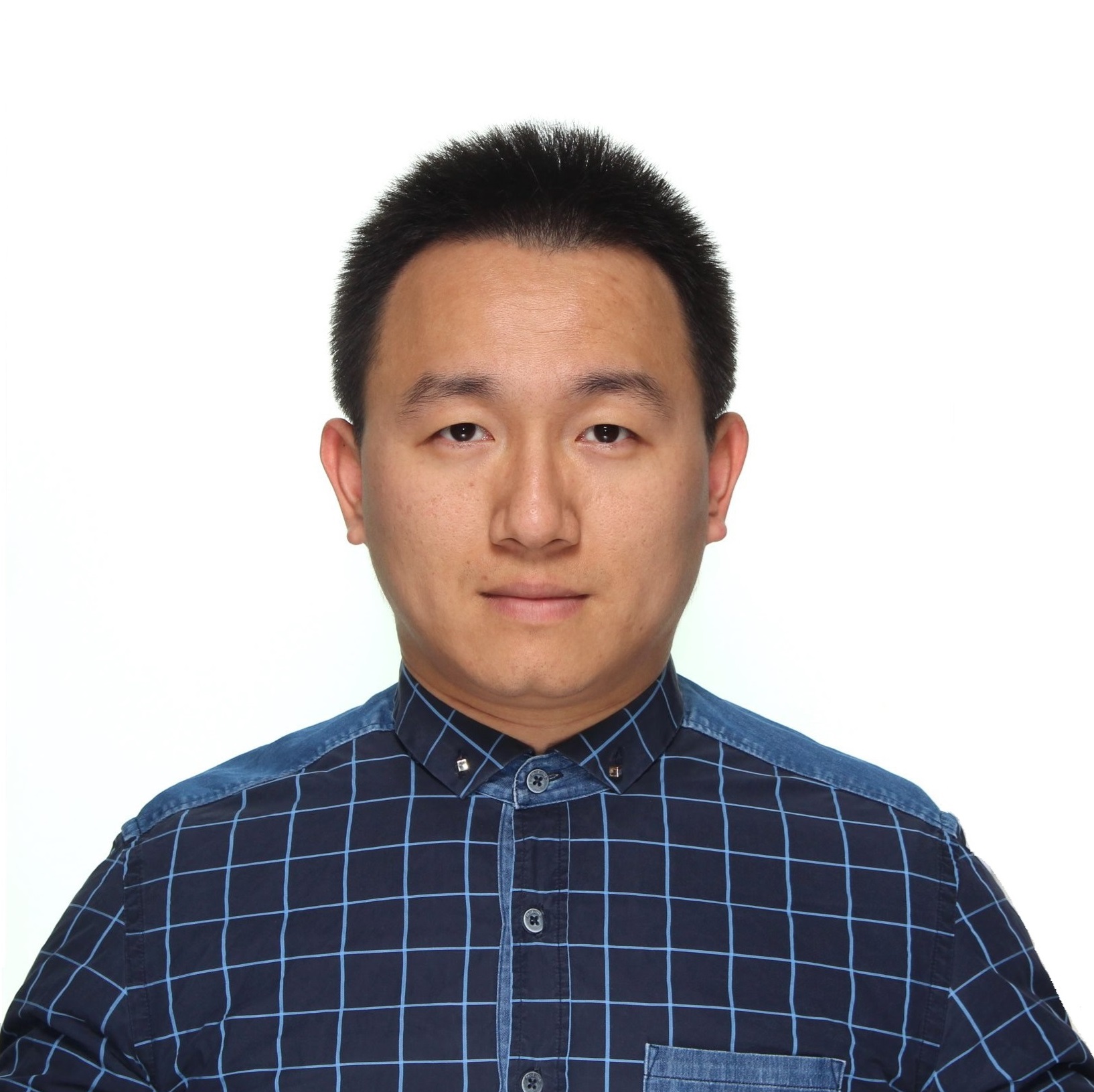}}]{Jiankang Deng} 
is a Lecturer at the Department of Computing, Imperial College London. He obtained his Ph.D. degree from Imperial College London in 2020. His research explores (1) multimodal foundation models, focusing on perceiving, understanding, modeling, and replicating complex human intelligence, and (2) generative modeling of the physical world, aiming at synthesizing scalable and reliable digital twins of real-world entities. His research lies at the intersection of computer vision and practical applications. He has more than 17K+ citations for his research and is one of the main contributors to the widely used open-source platform Insightface. He is an active area chair of prestigious computer vision and machine learning conferences (e.g., CVPR, ICCV, ICML, NeurIPS and ICLR). He is also an Associate Editor of IEEE Transactions on Image Processing. He is a Member of the IEEE.
\end{IEEEbiography}

\begin{IEEEbiography}[{\includegraphics[width=1in,,height=1.25in, clip,keepaspectratio]{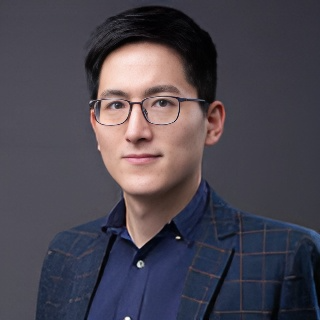}}]{Shan Luo} is a Reader (Associate Professor) at the Department of Engineering, King’s College London. Previously, he was a Lecturer at the University of Liverpool, and Research Fellow at Harvard University and University of Leeds. He was also a Visiting Scientist at the Computer Science and Artificial Intelligence Laboratory (CSAIL), MIT. He received the B.Eng. degree in Automatic Control from China University of Petroleum, Qingdao, China, in 2012. He was awarded the Ph.D. degree in Robotics from King’s College London, UK, in 2016. His research interests include tactile sensing, robot learning and robot visual-tactile perception. He is a Senior Member of the IEEE.
\end{IEEEbiography}

\end{document}